\documentclass[10pt,twocolumn,letterpaper]{article}

\usepackage{cvpr}              % To produce the CAMERA-READY version
\usepackage{algorithm}
\usepackage[accsupp]{axessibility}
\usepackage{amsmath}	
\usepackage{amssymb}	
\usepackage{booktabs}
\usepackage{dashrule}
\usepackage{microtype}
\usepackage{caption}
\usepackage{placeins}
\usepackage{color, colortbl} % Never used in text, thus allowed
\usepackage{siunitx}
\usepackage{enumitem}
\usepackage{tabularx}
\usepackage{xstring}
\usepackage{multirow}
\usepackage{nicefrac}
\usepackage{relsize}
\usepackage{xspace}
\usepackage{subcaption}
\usepackage{xcolor}
\usepackage[per-mode=fraction, fraction-command=\nicefrac]{siunitx}
\usepackage[super]{nth}

\DeclareSIUnit{\nothing}{\relax}
\DeclareSIUnit{\images}{img}

\definecolor{cvprblue}{rgb}{0.21,0.49,0.74}
\usepackage[pagebackref,breaklinks,colorlinks,allcolors=cvprblue]{hyperref}

\usepackage[capitalize,noabbrev]{cleveref}

\newcommand{\bh}{\mathbf{h}}

\definecolor{pinklayer}{HTML}{EC3C70}
\definecolor{yellowdinosaur}{HTML}{FCD267}
\definecolor{bluespot}{HTML}{1D91B8}
\definecolor{maskfusionblue}{HTML}{007191}
\definecolor{HMorange}{HTML}{F47A00}

\crefname{section}{Sec.}{Secs.}
\crefname{table}{Tab.}{Tabs.}
\crefname{figure}{Fig.}{Figs.}

\makeatletter
\DeclareRobustCommand\onedot{\futurelet\@let@token\@onedot}
\def\@onedot{\ifx\@let@token.\else.\null\fi\xspace}

\newcommand{\ourmethod}{MUFASA\xspace}
\newcommand{\ourdino}{DINOSAUR-M\xspace}
\newcommand{\ourspot}{SPOT-M\xspace}
\newcommand{\best}[1]{\textbf{#1}}
\newcommand{\secondbest}[1]{\underline{#1}}

\newcommand*{\myparagraph}[1]{\smallskip\noindent\textbf{#1}\hspace{0.2em}}
\newcommand*{\myparagraphnospace}[1]{\smallskip\noindent\textbf{#1}}
\def\eg{\emph{e.g}\onedot} 
\def\ie{\emph{i.e}\onedot} 
\def\cf{\emph{cf}\onedot} 
 \def\vs{\emph{vs}\onedot}
\def\wrt{w.r.t\onedot}

\Crefname{section}{Sec.}{Secs.}
\Crefname{table}{Tab.}{Tabs.}
\Crefname{figure}{Fig.}{Figs.}

\newcommand{\bs}{\mathbf{s}}

\newcommand{\bx}{\mathbf{x}}

\newcommand{\cH}{\mathcal{H}}

\newcommand{\cS}{\mathcal{S}}

\newcommand{\cU}{\mathcal{U}}

\newcommand{\nodata}{\text{--}}

\newcommand{\showdecrease}[1]{\makebox[0pt][l]{\fontsize{4.2pt}{4pt}\selectfont($\downarrow$#1\%)}}
\newcommand{\spm}[1]{\makebox[0pt][l]{\textsmaller[1]{\!{ $\pm$}#1}}}
\newcommand{\ours}{\textsmaller[1]{(ours)}}

\def\paperID{40087}
\def\confName{CVPR}
\def\confYear{2026}

\title{MUFASA: A Multi-Layer Framework for Slot Attention}

\newcommand{\AuthorBlock}{
Sebastian Bock\textsuperscript{\normalfont{}* 1,2} \quad Leonie Sch{\"u}{\ss}ler\textsuperscript{\normalfont{}* 1,2} \\
Krishnakant Singh\textsuperscript{\normalfont{} 1} \quad Simone Schaub-Meyer\textsuperscript{\normalfont{} 1,3} \quad Stefan Roth\textsuperscript{\normalfont{} 1,2,3} \\
{\normalsize \textsuperscript{1}TU Darmstadt \quad \textsuperscript{2}Zuse School ELIZA \quad \textsuperscript{3}hessian.AI \quad \textsuperscript{*}equal contribution}
}
\author{\AuthorBlock}

\begin{document}
\maketitle
\begin{abstract}
Unsupervised object-centric learning (OCL) decomposes visual scenes into distinct entities. Slot attention is a popular approach that represents individual objects as latent vectors, called slots. Current methods obtain these slot representations solely from the last layer of a pre-trained vision transformer (ViT), ignoring valuable, semantically rich information encoded across the other layers. To better utilize this latent semantic information, we introduce \textbf{MUFASA}, a lightweight plug-and-play framework for slot-attention-based approaches to unsupervised object segmentation. Our model computes slot attention across multiple feature layers of the ViT encoder, fully leveraging their semantic richness. We propose a fusion strategy to aggregate slots obtained on multiple layers into a unified object-centric representation. Integrating MUFASA into existing OCL methods improves their segmentation results across multiple datasets, setting a new state of the art while simultaneously improving training convergence with only minor inference overhead.\footnote{Project page \& code: \href{https://visinf.github.io/mufasa/}{https://visinf.github.io/mufasa/}}
\end{abstract}
    
\section{Introduction}
\begin{figure}[t]
    \centering
    \includegraphics[width=1\linewidth]{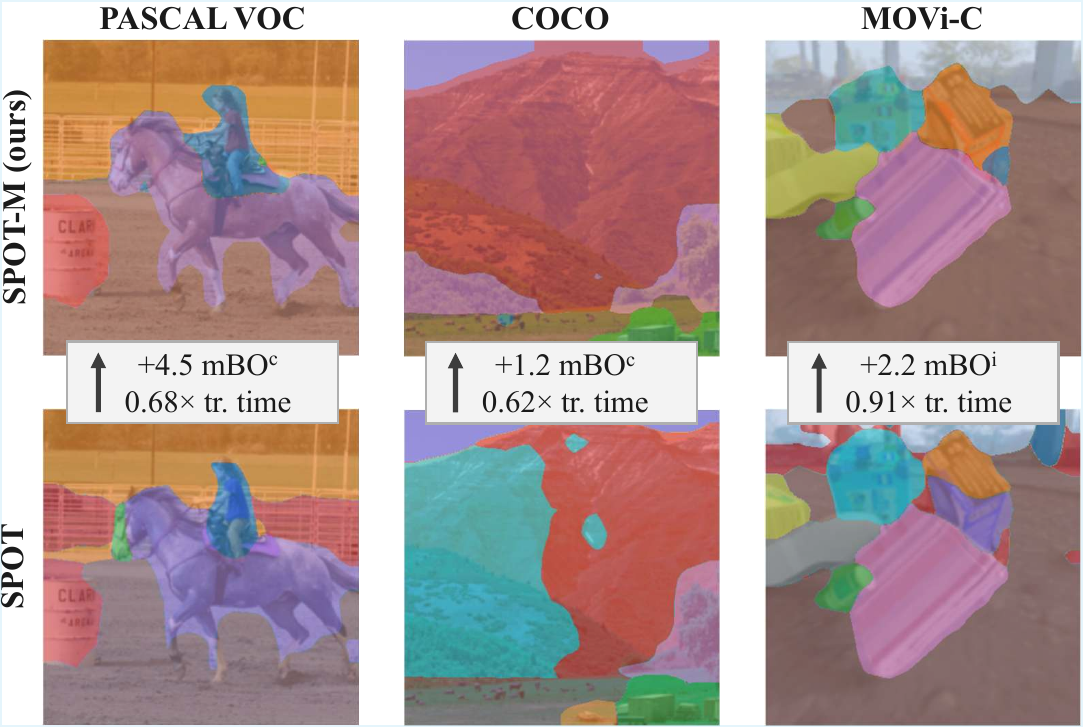}
    \vspace{-1.5em}
    \caption{\textbf{\ourmethod}. Our novel framework for slot-based methods leverages multiple feature layers of vision transformers for object-centric learning. Integrated into the current best model, SPOT \cite{kakogeorgiou2024spot}, we achieve a new state of the art in unsupervised object segmentation on PASCAL VOC, COCO, and MOVi-C, producing high-quality segmentation masks while requiring less time to train.
    }
    \label{intro:teaser}
    \vspace{-0.5em}
\end{figure}
\label{sec:introduction}
Object-centric learning (OCL) aims to decompose a scene into a set of object-specific representations in an unsupervised manner \cite{dittadi2021generalization}. This assumption is rooted in principles from human perception, suggesting that our visual system naturally segments a scene into meaningful entities \cite{kimch2015theperception}.
%Unlike more traditional architectures such as convolutional neural networks (CNNs), which produce a single representation for an entire scene, or vision transformers (ViTs) \cite{dosovitskiy2020image}, which have multiple tokens without specifically tying them to objects, object-centric representations explicitly model individual objects.
OCL methods have been used in various domains, ranging from building world models \cite{wuslotformer,collu2024slot}, robotics \cite{heravi2023visuomotor,liu2023structdiffusion}, explainability \cite{li2021scouter}, and compositional learning \cite{wu2023slotdiffusion,jiang2023object} to unsupervised object segmentation (UOS) \cite{kakogeorgiou2024spot,seitzer2022bridging}. Among various OCL approaches \cite{greff2019multi,burgess2019monet}, slot-attention (SA) methods have seen widespread adoption for their effectiveness in the unsupervised decomposition of scenes into objects \cite{locatello2020object}. Here, input features are grouped into a set of latent vectors, termed slots, and iteratively refined through an attention-based mechanism, where slots compete to bind to individual objects. Initially, most applications remained limited to synthetic and constrained datasets \cite{johnson2017clevr,groth2018shapestacks}. DINOSAUR \cite{seitzer2022bridging} scaled slot-attention-based methods to real-world datasets by utilizing a pre-trained DINO \cite{caron2020unsupervised} encoder for feature reconstruction. Subsequent works leveraged teacher-student architectures to guide slot binding \cite{kim2024bootstrapping,didolkar2025ftdinosaur}, including SPOT \cite{kakogeorgiou2024spot}, establishing a new state of the art (SOTA) in UOS.

Both DINOSAUR and SPOT utilize features extracted from a pre-trained DINO ViT \cite{caron2021emerging} for slot attention. As shown by \cite{amir2023on}, early layers of DINO ViTs capture positional information, while semantic content emerges in middle layers and becomes increasingly rich until the final layer. Thus, semantically meaningful features are not confined to the final layer. Instead, valuable information is present across several layers, which encode complementary semantics. Consequently, restricting the input of slot attention to the last encoder layer does not leverage all semantic information offered by the DINO ViT. To address this limitation, we propose \textbf{MUFASA}, a \textbf{Mu}lti-Layer \textbf{F}r\textbf{a}mework for \textbf{S}lot \textbf{A}ttention, as a novel and lightweight plug-and-play framework for slot-attention models utilizing DINO features. We leverage the rich semantics encoded across several layers \cite{amir2023on,yao2024denseconnector} by simultaneously using multiple encoder layers for slot attention. The slots emerging from multiple slot-attention modules are aligned in terms of their object information using Hungarian matching \cite{kuhn1955hungarian}. The matched slots are fed into a fusion module that integrates the slots from different layers into a unified representation before passing them to the decoder. Our multi-layer method learns to segment objects across multiple feature representations; utilizing this diversity allows \ourmethod to better segment objects (\cref{intro:teaser}). 
We integrate \ourmethod into DINOSAUR and SPOT, substantially improving their segmentation quality. With this, we set a new SOTA on the VOC, MOVi-C, and COCO datasets (\cf \cref{intro:teaser}) while simultaneously reducing training times.

In summary, our contributions include: \emph{(i)} We propose a novel slot-attention framework leveraging the complementary feature representations of multiple DINO layers for unsupervised object segmentation. Our framework includes M-Fusion, a technique to effectively combine multi-layer slots into a unified representation.
 \emph{(ii)} \ourmethod is plug-and-play, enabling simple integration into slot-attention models utilizing DINO encoders.
 \emph{(iii)} Applying \ourmethod improves previous OCL methods for unsupervised object segmentation in nearly all settings.
 \emph{(iv)} \ourmethod is lightweight: With minimal parameter and inference overhead, faster training efficiency is achieved.
 \emph{(v)} Integrated into SPOT, we achieve new SOTA results on COCO, PASCAL VOC, and MOVi-C.

 %\emph{(v)} Our approach maintains strong performance across a variety of encoder and decoder architectures and layer configurations, highlighting its robustness and generalizability. 

\section{Related Work}
\label{sec:relatedwork}

\myparagraph{Object-centric learning.}
Early methods for OCL utilized sequential architectures \cite{burgess2019monet,greff2019multi,jiang2019scalor,lin2020space,eslami2016attend,li2020learning,lin2020improving}. These methods do not scale well to complex scenes % due to their sequential nature, 
and impose an arbitrary order on the objects in a scene.
To resolve these issues, \cite{engelcke2020genesis} formulated OCL as an instance coloring problem and used a stick-breaking prior. Another line of work, called slot attention (SA), uses a soft $k$-means clustering approach, wherein object latents are learned iteratively via clustering of similar features \cite{locatello2020object}. Cluster separation is achieved by applying dot-product attention with softmax over cluster centers, \ie slots. 
SA, until recently, only showed promising results on synthetic and constrained datasets \cite{johnson2017clevr,karazija2021clevrtex,groth2018shapestacks}. DINOSAUR~\cite{seitzer2022bridging} showed that it is possible to scale SA-based methods to complex, real-world scenes \cite{lin2014microsoft,everingham2010pascal,Geiger2013IJRR} by learning in the feature space of a pre-trained self-supervised encoder \cite{oquab2023dinov2,caron2020unsupervised} instead of on raw pixels. Since then, SA methods have seen a resurgence of interest, with many works~\cite{Gong_2025_ICCV} building upon \cite{seitzer2022bridging}. 
In \cite{pramanik2024masked}, a masking scheme discards background features, and multi-query SA processes the same feature layer with multiple independent SA modules in an ensemble-like fashion.
\cite{kakogeorgiou2024spot} introduced SPOT, which leverages attention-based self-training to distill knowledge from a teacher to a student model via a cross-entropy loss between their attention masks. They further proposed patch-order permutations within autoregressive transformer decoders, altering the reconstruction order of image patches during decoding, achieving new SOTA results in UOS.

\myparagraph{Decoders for slot attention.}
A key component of the SA architecture is the decoder. Early work \cite{eslami2016attend,kosiorek2018sequential} used a patch decoder. Alternatively, \cite{kipf2022conditional,locatello2020object} used a spatial-broadcast decoder \cite{watters2019spatial} to predict RGB images and segmentation masks from each slot, which are then combined through alpha masking.  
However, this per-slot decoding strategy limits the application to synthetic datasets \cite{groth2018shapestacks,johnson2017clevr,karazija2021clevrtex}.
SLATE~\cite{singh2021illiterate} and STEVE \cite{singh2022simple} proposed the use of powerful transformer models \cite{vaswani2017attention} as decoders. They use a dVAE \cite{im2017denoising} to tokenize the input, and
train the slot-conditioned transformer decoder to autoregressively reconstruct patch tokens, enabling to segment more complex images.
\cite{seitzer2022bridging} noted that while an MLP decoder separates instances better during reconstruction, a transformer decoder is generally more expressive and produces tighter segmentation masks with cleaner background segmentation. 
%However, as noted by \cite{seitzer2022bridging}, using a transformer decoder for feature reconstruction biases the model to decompose the scene into semantic classes, causing under-segmentation issues, whereas a simple MLP decoder helps decompose the scene into distinct objects. 
%Also, transformer-based decoders can perform worse than a CNN-based decoder for image reconstruction \cite{singh2021illiterate,wuslotformer}. 
%Thus, we report our results on both the transformer- and MLP-based decoders. 
Another line of work focuses on SA for compositional generation \cite{jiang2023object,wu2023slotdiffusion, singh2025glass}, and shows that using a pre-trained diffusion decoder can improve compositional generation. However, most of these methods are inferior to feature reconstruction methods such as SPOT in UOS.

\myparagraph{Multi-layer approaches.} %have long been a cornerstone in computer vision \cite{epshtein2005feature,krishnapuram1992fuzzy,lades1993distortion,cantoni1995hierarchical,chen2020towards}. 
Vision transformers, particularly DINO ViTs \cite{caron2021emerging}, have been shown to learn a representational hierarchy across layers \cite{amir2023on}. Here, shallow layers mainly contain spatial information, while semantics emerge in intermediate layers and become increasingly rich in deeper layers. This hierarchy is distinct from the scale-based hierarchies in CNNs, which typically follow a coarse-to-fine progression over spatial resolutions \cite{zeiler2013visualizing}. Instead, ViTs exhibit a more uniform representation across all layers \cite{raghu2022visiontransformerslikeconvolutional}. Yet, the individual layers exhibit distinct layer-wise behavior in downstream tasks, indicating semantically complementary encodings \cite{vanyan2024analyzinglocalrepresentationsselfsupervised}. Recently, several methods showed the potential of leveraging features from multiple ViT layers for multi-modal tasks \cite{yao2024denseconnector,cao2024mmfuser}, feature forecasting~\cite{karypidis2025dinoforesight}, visual correspondence~\cite{zhang2023tale, cho2021cats}, and object discovery~\cite{Lin_2023_WACV}. However, the integration of multi-layer ViT representations into slot-attention-based methods remains unexplored to date.  

%\myparagraph{Hierarchical feature representations} have been a cornerstone of computer vision \cite{epshtein2005feature,krishnapuram1992fuzzy,lades1993distortion,cantoni1995hierarchical,chen2020towards}. In CNNs, this hierarchy corresponds to a coarse-to-fine progression over spatial scales: early layers capture low-level features, while later layers encode increasingly complex and semantic concepts, which has proven effective in many tasks \cite{ren2015faster,redmon2016yolo,girshick2014rich}. Vision Transformers, especially DINO-ViTs, have also been shown to learn hierarchical representations across layers \cite{amir2023on}. While CNNs rely on multi-scale features, ViTs exhibit a different representational hierarchy where shallow layers mainly contain spatial information, and semantics emerge in intermediate layers and become increasingly rich in deeper layers. Recently, several methods \cite{yao2024denseconnector,cao2024mmfuser} show the potential of leveraging features from multiple ViT layers for various multi-modal tasks. However, the integration of multi-layer ViT representations into slot attention-based methods remains unexplored to date.
 
\section{MUFASA}
\label{sec:method}

\subsection{Preliminaries}
\paragraph{Autoencoding slot-attention (SA) architecture.} SA-based methods commonly use an encoder-decoder architecture with an SA bottleneck. The encoder extracts $N$ patch-wise features $\bh \in \mathbb{R}^{N \times d_\mathrm{emb}}$ of dimensionality $d_\mathrm{emb}$ from an input image $\bx \in \mathbb{R}^{H \times W \times C}$. Typically, a self-supervised, pre-trained ViT is used as encoder, \eg, DINO \cite{caron2021emerging}. A subsequent SA module groups these features into a set of $K \ll N$ latent vectors $\cS = \{\bs_k\in\mathbb{R}^{d_\mathrm{slot}} \mid k=1,\ldots,K\}$%
%$\cS \in \mathbb{R}^{K \times d_\mathrm{slot}}$
, where each item $\bs_k$ 
%with indices $k \in \{1, ...,K\}$
-- called a slot -- is of dimensionality $d_\mathrm{slot}$. 
Given these slots, the decoder network reconstructs the input signal from the slots. The whole architecture is trained end-to-end using a normalized reconstruction loss between the decoder output and the encoder feature representations \cite{seitzer2022bridging}:

\begin{equation}
\label{eq:rec_loss}
\mathcal{L}_{\mathrm{Rec}} = \frac{1}{N \cdot d_\mathrm{emb}} \big\lVert \bh - \mathrm{Decoder}(\cS) \big\rVert_2^2.
\end{equation}

% Following prior work \cite{seitzer2022bridging, kakogeorgiou2024spot}, we use features from a self-supervised pretrained DINO-ViT encoder \cite{caron2021emerging}.
% To evaluate the quality of the slots in this unsupervised setting, we can examine the slot attention masks $A \in \mathbb{R}^{d_h \times k}$ associated with the slots as segmentation masks. The slot attention masks represent the degree to which each slot attends to each image patch, allowing us to assign every image patch to the slot that attended most to it. 
% Slot attention masks can be derived from two locations, the SA module and the transformer decoder used for reconstruction.

% This must be said again: Slot attention masks can be derived from two locations, the SA module and the transformer decoder used for reconstruction.

%To reconstruct these features from the latent slots, an auto-regressive transformer decoder design is used, based on \cite{seitzer2022bridging}. This architecture generates the output sequentially patch by patch based on the slot representations, predicting each token by conditioning on the previous ones.
\myparagraphnospace{Slot attention}
performs an iterative refinement that maps the set of input features $\bh$ to the set $\cS$ of $K$ output slots. Initially, the slots are independently sampled from a Gaussian distribution. Then, the slots are iteratively updated by computing the dot-product attention \cite{vaswani2017attention} between the input features and the slots from the previous iteration. Here, learned linear transformations (LLT), $f_{\mathrm{Key}}$ and $f_{\mathrm{Query}}$, map the patch-wise input features onto keys and the slots onto queries in a common $d$-dimensional space. Attention scores are then obtained via a scaled-dot product and softmax normalization \cite{vaswani2017attention}, yielding a probability distribution over slots for each input feature. This enforces a competition between the slots to bind to meaningful areas in the image. This yields the slot-attention matrix $\mathcal{A}^{\mathrm{Slot}} \in \mathbb{R}^{N\times K}$, which denotes the assignment of each image patch (token) to the $K$ slots:
\begin{equation}
\label{eq:dot_product_attention}
    \mathcal{A}^{\mathrm{Slot}} = 
    \underset{K}{\mathrm{softmax}}\left(\frac{f_{\mathrm{Key}}(\bh) \cdot f_{\mathrm{Query}}(\cS)^T}{\sqrt{d}}\right).
\end{equation}
The slot updates are computed as a weighted aggregation of input features with $\mathcal{A}^{\mathrm{Slot}}$. Finally, the slots are iteratively updated using a learned recurrent function \cite{cho2014learning}.

\myparagraph{Decoder.} Following \cite{singh2021illiterate}, we use an auto-regressive transformer decoder to reconstruct the output sequentially patch by patch based on the slots, predicting each token by conditioning on the previous ones. The decoder includes multiple patch-to-slot cross-attention layers to enable slots to guide reconstruction. Thus, attention masks denoting how slots attend to different image patches can also be obtained from the final self-attention layer of the decoder. For this, the slots $\cS$ are mapped to keys using an LLT $g_{\mathrm{Key}}$, while the reconstructed image patches $\mathbf{y}_\mathrm{prev}$ of the previous attention layer are mapped to queries in the same $d$-dimensional space utilizing the LLT $g_{\mathrm{Query}}$. Computing the dot product between the keys and queries with subsequent softmax normalization yields an attention mask at every attention head. Averaging over all $J$ heads produces the final decoder attention mask:
\begin{equation}
    \small
    \mathcal{A}^{\mathrm{Dec}} = \frac{1}{J} \sum_{j=1}^J\underset{K}{\mathrm{softmax}}\left(\frac{g_{\mathrm{Query}}(\mathbf{y}_\mathrm{prev})_j\cdot g_{\mathrm{Key}}(\cS)_j^T}{\sqrt{d}}\right).
    %\raisetag{38pt}
\end{equation}
A segmentation mask can be derived from the attention maps of either the slot-attention module or the decoder. By applying an $\mathrm{argmax}$ operation across the slot dimension, we assign each image patch to the slot that attended most to it.
\begin{figure}[t!]
    \centering

    % Resize only the original grid part
    \resizebox{\columnwidth}{!}{
    \begin{tabular}{@{}c@{\hskip 0.4in}c@{\hskip 0.04in}c@{\hskip 0.04in}cc@{}}%\hskip 0.04in}} 

        % First row
        \begin{subfigure}{0.25\textwidth}
            \centering
            \includegraphics[width=\textwidth]{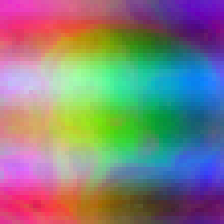}
        \end{subfigure} &
        \begin{subfigure}{0.25\textwidth}
            \centering
            \includegraphics[width=\textwidth]{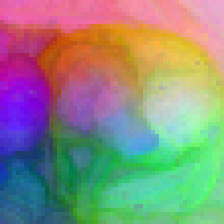}
        \end{subfigure} &
        \begin{subfigure}{0.25\textwidth}
            \centering
            \includegraphics[width=\textwidth]{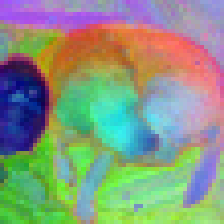}
        \end{subfigure} &
        \begin{subfigure}{0.25\textwidth}
            \centering
            \includegraphics[width=\textwidth]{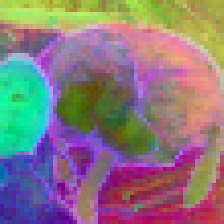}
        \end{subfigure} &
        \hspace{2em}
        \begin{subfigure}{0.25\textwidth}
            \centering
            \includegraphics[width=\textwidth]{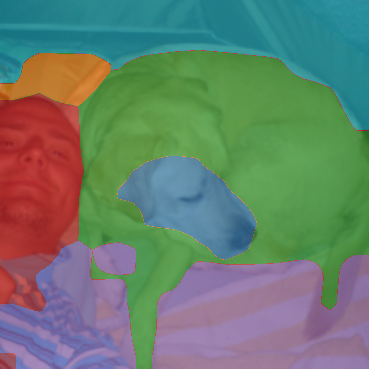}
        \end{subfigure} \\[1mm]

        % Caption for first row
        \multicolumn{4}{c}{\huge (a) PCA of DINO features at layers 4, 10, 11 and 12 \vspace{0.4em}} & 
        \multicolumn{1}{c}{\huge (c) Single-layer model}  \vspace{0.5em} \\

        % Second row
        \begin{subfigure}{0.25\textwidth}
            \centering
            \includegraphics[width=\textwidth]{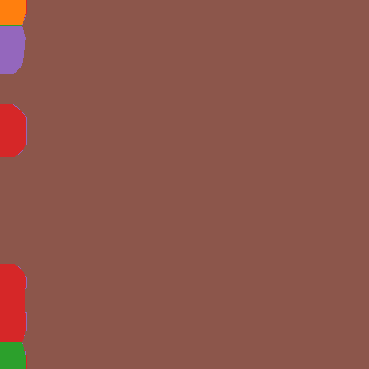}
        \end{subfigure} &
        \begin{subfigure}{0.25\textwidth}
            \centering
            \includegraphics[width=\textwidth]{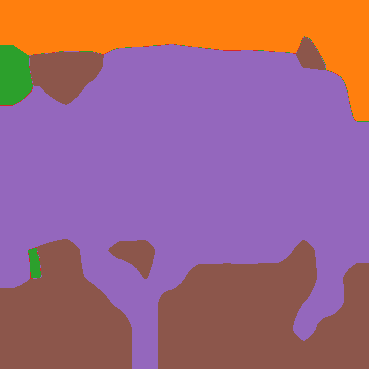}
        \end{subfigure} &
        \begin{subfigure}{0.25\textwidth}
            \centering
            \includegraphics[width=\textwidth]{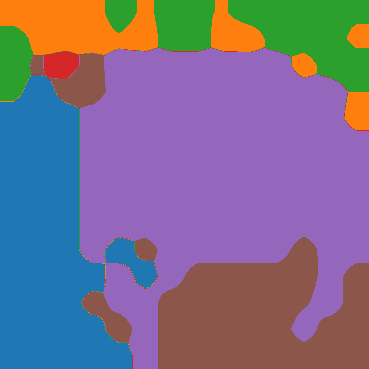}
        \end{subfigure} &
        \begin{subfigure}{0.25\textwidth}
            \centering
            \includegraphics[width=\textwidth]{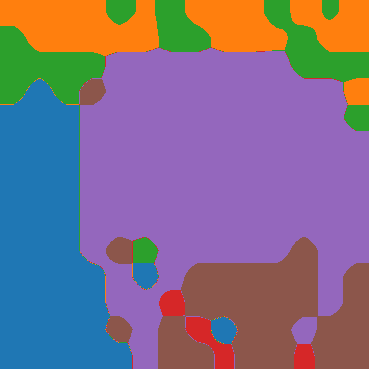}
        \end{subfigure} &
        \hspace{2em}
        \begin{subfigure}{0.25\textwidth}
            \centering
            \includegraphics[width=\textwidth]{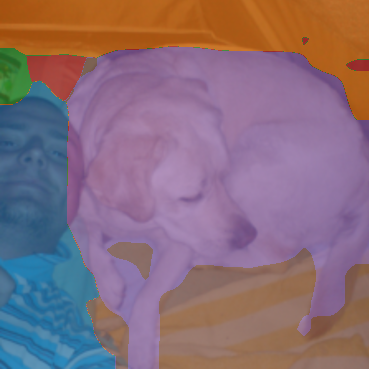}
        \end{subfigure} \\[1mm]

        % Caption for second row
        \multicolumn{4}{c}{\huge (b) Slot-attention masks per layer} \vspace{0.5em}& 
        \multicolumn{1}{c}{\huge (d) Fused SA mask} 
         \vspace{0.5em} \\

    \end{tabular}
    }
    %\begin{picture}(0,0)
    %\put(-115,-10){\scriptsize(e)}  % X and Y offsets
    %\end{picture}
    \includegraphics[width=1\columnwidth]%
    {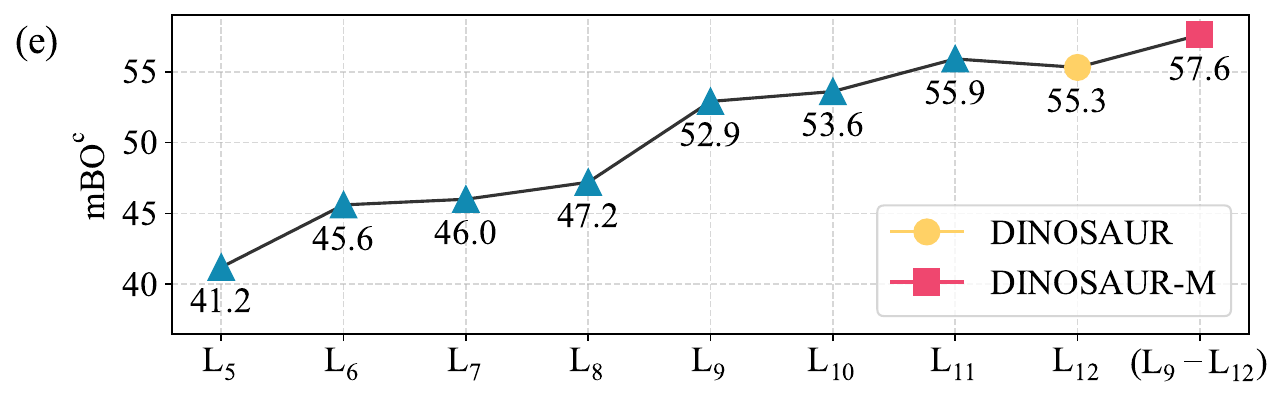}
    \vspace{-1.5em}
    \caption{\textbf{Complementarity of DINO layers.} \emph{(a)} PCA visualization for features from layers 4 and 10--12, each encoding varying semantics. \emph{(b)} Corresponding attention masks from slot attention on these layers, showing different segmentations.  \emph{(c)} Segmentation mask of the single-layer SPOT. \emph{(d)} The fused slot-attention mask of our \ourspot captures the person and the dog in a single slot each and follows their boundaries more closely. \emph{(e)} Gain by combining layers. \textcolor{bluespot}{Blue} shows the segmentation accuracy of single-layer DINOSAUR models trained on different encoder layers, \textcolor{yellowdinosaur}{yellow} is the original DINOSAUR using $\textrm{L}_{12}$. \textcolor{pinklayer}{MUFASA} on DINOSAUR combines multiple layers, surpassing all individual ones.}
    %\textcolor{pinklayer}{Red} bars show DINOSAUR with different encoder layers; \textcolor{yellowdinosaur}{yellow} shows the original. \textcolor{bluespot}{MUFASA} on top of DINOSAUR combines multiple layers, which surpasses all individual ones.}
    \label{fig:hierarchy_with_SA}
    \vspace{-0.6em}
\end{figure}

\subsection{Multi-layer slot attention}
\begin{figure*}
    \centering
    \includegraphics[width=1\linewidth]{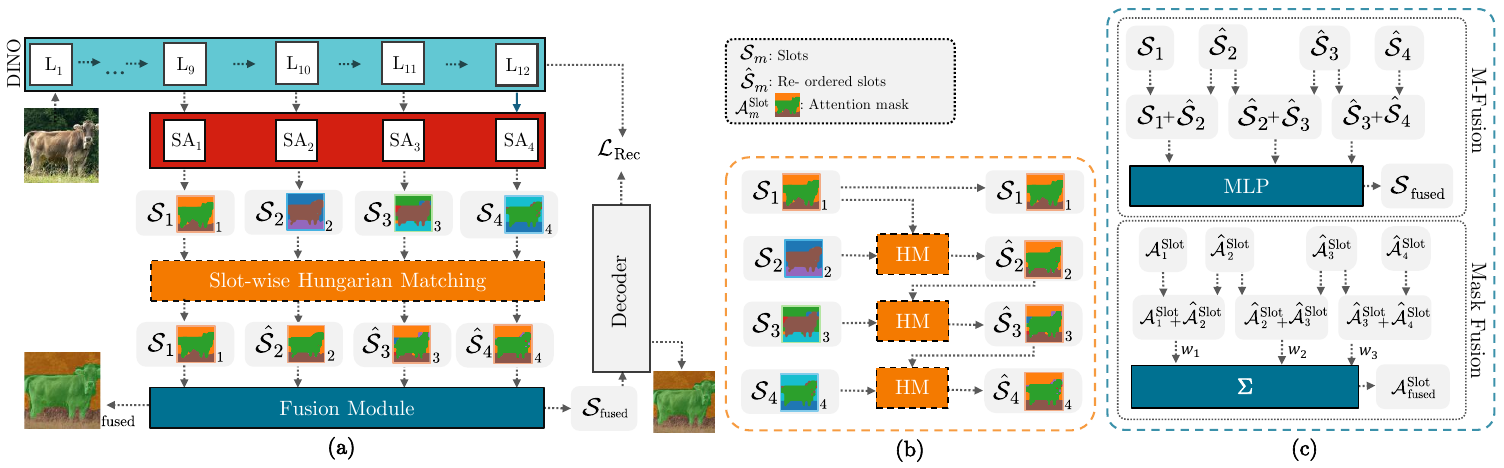}
    \vspace{-1.5em}
    \caption
    {\textbf{MUFASA architecture.} \emph{(a)} For an input image, features from multiple layers of a DINO encoder are processed by multiple slot-attention (SA) modules, each producing slots $\cS_m$ and corresponding attention masks $\mathcal{A}_m^{\mathrm{Slot}}$. After Hungarian matching, a fusion module merges slots and masks. A ViT decoder reconstructs the last encoder layer’s features from fused slots, yielding the decoder attention mask $\mathcal{A}^{\mathrm{Dec}}$. The reconstruction loss $\mathcal{L}_{\mathrm{Rec}}$ guides training. \textbf{\emph{(b)} \textcolor{HMorange}{Hungarian matching (HM)}}. The set of slots and attention masks are re-ordered for best correspondence across layers. \textbf{\emph{(c)} \textcolor{maskfusionblue}{Fusion module}}. The re-ordered set of slots and masks are summed in adjacent pairs. Slots are projected into a fused representation $\cS_{\mathrm{fused}}$, while a weighted combination of attention masks produces the fused mask $\mathcal{A}_{\mathrm{fused}}^{\mathrm{Slot}}$.
    }
    \label{fig:architecture_diagram}
    \vspace{-0.5em}
\end{figure*}

Existing methods in unsupervised object segmentation using slot attention typically  use solely the features from the final layer of a pre-trained encoder, \eg, DINO \cite{seitzer2022bridging, kakogeorgiou2024spot}. However, semantic information is not confined to the final layer \cite{amir2023on}. As shown in \cref{fig:hierarchy_with_SA}~(e), the feature representations obtained at various layers of a DINO encoder -- particularly deeper ones -- enable a strong segmentation capability when utilized as input and reconstruction target in slot attention. This is further supported by \cref{fig:hierarchy_with_SA}~(a), where PCA decompositions of features from intermediate deep layers reveal semantically meaningful but distinct spatial patterns. These differences manifest in the resulting slot-attention masks, which vary across layers (\cf \cref{fig:hierarchy_with_SA}~(b)). In contrast, early layers (\eg, layer 4) exhibit coarse structures in the PCA, which yield insufficient segmentation masks. By combining slots from those layers that individually yield the best segmentations, we obtain a fused representation that produces more accurate segmentations than any individual layer alone (\cf \cref{fig:hierarchy_with_SA}~(e)). For instance, as shown in \cref{fig:hierarchy_with_SA}~(d), the fused mask more precisely outlines the person and dog and further reduces over-segmentation (\eg, the dogs' face) and missing details (\eg, the paws) observed in the SPOT model (\cref{fig:hierarchy_with_SA}~(c)). These improvements arise from complementary information encoded across layers: $\textrm{L}_{10}$ merges the dog and the person into one slot, while later layers correctly separate them but introduce background noise, which is mitigated in the fused mask. This suggests that leveraging the semantic richness of intermediate ViT layers can benefit object-centric learning. Accordingly, we propose a multi-layer slot-attention framework, MUFASA (\cref{fig:architecture_diagram}), that integrates features from multiple encoder layers to enhance UOS.

\myparagraph{Integration of \ourmethod.} We design \ourmethod as a simple plug-and-play component, allowing seamless integration into existing slot-attention-based methods relying on a pre-trained DINO encoder. To this end, feature representations are extracted at multiple layers of the encoder and the single-layer SA bottleneck of the model is replaced with our proposed multi-layer SA module. Integrating our approach into DINOSAUR and SPOT yields \ourdino and \ourspot, respectively. \ourmethod is trained with no additional losses, solely utilizing training signals of its respective base model.

\myparagraph{Multiple feature layers.} 
The pre-trained DINO-ViT encoder produces a set of feature representations $\cH = \{\bh_1, \bh_2, .., \bh_{12}\}$ across its 12 layers. Instead of extracting only the final representation $\bh_{12}$, we define an index set $\mathcal{I} \subseteq \{1, \ldots, 12\}$ and use it to select a subset $\hat{\cH}$ of $\cH$:
\begin{equation}
\label{subsetH}
\hat{\cH} \subseteq \cH, \quad \hat{\cH} = \left\{ \bh_i \in \mathbb{R}^{N \times d_\mathrm{emb}} \mid i \in \mathcal{I}\right\},
\end{equation}
where, as above, $N$ denotes the number of tokens and $d_\mathrm{emb}$ is the feature dimension of the tokens. We restrict the index set size $|\mathcal{I}|$ to some value $M$. Given the subset of feature vectors $\hat{\cH}$, we perform slot attention on every single $\bh_i \in \hat{\cH}$. 
%The sets $\hat{\mathcal{H}}$ and $\mathcal{U}$ consequently have the same number of elements, so that $h$ at position $i$ in $\hat{\mathcal{H}}$ corresponds to $\cS$ at position $m$. 
With this, we obtain a family of $M$ slot sets $\cU =\{\cS_1, \cS_2, \ldots, \cS_{M}\}$. Each $\cS_{m}$, with indices $m \in \{ 1, \ldots, M\}$, consists of its own $K$ slot vectors $\bs_k^m$ with indices $k \in \{1, ...,K\}$, \ie $\cS_{m} = \{\bs_1^m, \ldots, \bs_K^m\}$. To obtain each $\cS_m$, we initialize an independent slot-attention module $\mathrm{SA}_m$ with its own set of trainable parameters, rather than sharing weights between them, to enable adaptation to layer-specific features and capture more diverse information. Consequently, we obtain a slot-attention mask $\mathcal{A}^{\mathrm{Slot}}_m$ for every $\mathrm{SA}_m$, which denotes the attendance of image patches within the corresponding feature level to the slots of $\cS_m$.

%To leverage the surplus of information, the sets of output slots have to be projected into a single set of slots that can be passed on to the decoder, following the intuition that this leads to an improved internal slot representation. Otherwise, one could argue for the use of multiple decoders instead of projecting the sets into a single representation. We have opted not to do this to keep the computational overhead minimal. Utilizing the fused slot representation, the decoder tries to reconstruct the patch-wise feature vectors given by the last encoder layer as a training signal.

\myparagraph{Slot fusion.} To enable the decoder to leverage the additional information, the family of slots $\cU$ is projected onto a single set of slots $\cS_{\mathrm{fused}} \in \mathbb{R}^{K\times d_\mathrm{slot}}$. This allows the semantic information encoded across layers to be integrated into a slot-based representation that can be utilized by an auto-regressive transformer decoder. We term this process \textit{slot fusion}. Prior to fusion, we ensure that two sets of slot vectors stemming from subsequent layer indices $\cS_m =\{\bs_1^m, \ldots, \bs_K^m\}$ and $\cS_{m+1} = \{\bs_1^{m+1}, \ldots, \bs_K^{m+1}\}$ are aligned in the sense that the slots $\bs^m_k$ and $\bs^{m+1}_k$ with corresponding indices
$k \in \{1, ...,K\}$ across layers learn to bind to the same object. This is achieved by computing a permutation $\Pi_{m+1}$ via Hungarian matching \cite{kuhn1955hungarian} based on maximizing the mean Intersection-over-Union ($\text{mIoU}$) metric between the corresponding binarized slot-attention masks $\mathcal{A}_{m}^{\mathrm{Slot}}$ and $\mathcal{A}_{m+1}^{\mathrm{Slot}}$ (see \cref{fig:architecture_diagram} (b)). The resulting maximum $\text{mIoU}$ assignment $\Pi_{m+1}$ allows us to reorder the indices of the set of slots of $\cS_{m+1}$% of $\mathcal{A}_{m+1}^{\mathrm{Slot}}$ to $\hat{\mathcal{A}}_{m+1}^{\mathrm{Slot}}$
, ensuring that the best matching slots are bound to the same indices across layers. We reorder $\cS_{m+1}$ and its attention mask $\mathcal{A}_{m+1}^{\mathrm{Slot}}$ based on $\Pi_{m+1}$, starting with $m=1$. This results in the aligned sets of slots $\hat{\mathcal{U}} = \{\hat{\cS_1},\hat{\cS}_2, \ldots, \hat{\cS}_M\}$, where $\hat{\cS_1}=\cS_1$, and corresponding masks $\hat{\mathcal{A}}_{m}^{\mathrm{Slot}}$ for indices $m\in\{1,\ldots,M\}$. 

We propose a novel approach to slot fusion, termed \textbf{M-Fusion} (\cref{fig:architecture_diagram} (c)), designed to capture non-linear relations between slots of multiple layers. At its core, the fusion is done via a learned projection of $\cU$ using a multi-layer perceptron (MLP). First, the $M$ aligned slot sets $\hat{\mathcal{U}} = \{\hat{\cS_1},\hat{\cS}_2, \ldots, \hat{\cS}_M\} $ have to be concatenated into a single set of slots. Inspired by \cite{yao2024denseconnector}, we take each subsequent pair of slot sets $(\hat{\cS}_m, \hat{\cS}_{m+1})$ in a sliding window-like fashion and sum corresponding slot vectors, effectively encoding an inductive bias of local interactions between adjacent slots. By this, we incorporate multi-layer information as features and obtain $M-1$ elements $\mathcal{Z}=\{(\hat{\cS}_1 + \hat{\cS}_2), \ldots , (\hat{\cS}_{M-1} + \hat{\cS}_M)\}$. After that, we concatenate $\mathcal{Z}$ along the slot feature dimension. An MLP projects this intermediate representation into a fused set of slots:
\begin{equation}
    \cS_{\mathrm{fused}} = \mathrm{MLP}\left(\mathrm{Concat}(\mathcal{Z}, \mathrm{axis = features})\right).
\end{equation}
The slot-attention masks $\hat{\mathcal{A}}_{m}^{\mathrm{Slot}}$, corresponding to the slots $\hat{\cS}_m$, must be fused to a joint representation $\mathcal{A}_{\mathrm{fused}}^{\mathrm{Slot}}$ as well (\textit{mask fusion}). Analogously to slots, we add each successive pair of slot-attention masks together, resulting in $\mathcal{Z}^{\mathrm{att}}=\{( \hat{\mathcal{A}}_{1}^{\mathrm{Slot}} + \hat{\mathcal{A}}_{2}^{\mathrm{Slot}}), \ldots , (\hat{\mathcal{A}}_{M-1}^{\mathrm{Slot}}+ \hat{\mathcal{A}}_{M}^{\mathrm{Slot}})\}$. 
The resulting attention masks are then fused using a weighted linear combination
\begin{equation}
\mathcal{A}_{\mathrm{fused}}^{\mathrm{Slot}} = \sum_{m=1}^{M-1} w_m \mathcal{Z}^{\mathrm{att}}_m.
\label{eq:attmapfusion}
\end{equation}
If no teacher-student training is employed (\ourdino), the mask fusion weights $w \in \mathbb{R}^{M-1}$ are set to a constant uniform value of $\frac{1}{M-1}$, giving equal importance to each layer pair. When self-training is used (\ourspot), the weights are learned during training, with the knowledge distillation from the slot-attention masks of the teacher to the masks of the student as guiding signal. The mask-fusion weights are normalized by applying a softmax over the layer dimension.

%\myparagraph{MUFASA as a Plug-and-Play Module.}  
%MUFASA can be seamlessly integrated into popular state-of-the-art OCL methods that rely on a pre-trained DINO encoder. For example, in \textsc{DINOSAUR}, we replace the original single-layer slot attention module with our proposed multi-layer slot attention framework (\textsc{MUFASA}). In this configuration, the attention masks are fused by taking their uniform average; that is, the fusion weights $w \in \mathbb{R}^{M-1}$ are set to $\frac{1}{M-1}$, as described in \cref{eq:attmapfusion}.
%In the case of \textsc{SPOT}, we apply a similar strategy during teacher training by mean-fusing the attention maps. However, during student training, we treat the fusion weights $w$ as learnable parameters. These are optimized via a cross-entropy loss between the fused attention maps of the teacher and those of the student, defined as:
%\begin{equation}
%\tiny
%\mathcal{L}_{\text{mask}} = \text{CE}\left(\sum_{i=1}^{M-1} w_i \cdot \mathchal{A}_i^{(\text{student})}, \sum_{i=1}^{M-1} \frac{1}{M-1} \cdot \mathcal{A}_i^{(\text{teacher})} \right)
%\end{equation}
%Here, $\text{CE}(\cdot,\cdot)$ denotes the cross-entropy loss, and $\mathcal{A}_i$ refers to the attention mask from the $i^\text{th}$ layer.
% If self-training is employed, we utilize the distillation loss to obtain $w$ as learnable parameters during training and normalize them through a softmax.

\section{Experiments}
\label{sec:experiments}

\paragraph{Datasets.} We conduct experiments on multiple datasets (real and synthetic) to assess the ability of our approach to perform unsupervised object segmentation (UOS). 
For real-world images, we utilize the PASCAL VOC \cite{everingham2010pascal} dataset, which contains one or few salient objects per image. Furthermore, we evaluate on the COCO \cite{lin2014microsoft} dataset, as it offers real-world scenes of higher complexity with objects of diverse classes per image. We also leverage the MOVi-C dataset generated by the Kubric simulator \cite{greff2022kubric} for synthetic images of multi-object scenes with realistic variations in appearance and arrangement. Since MOVi-C is originally video-based, the usage for object segmentation is enabled by sampling random frames as done by \cite{seitzer2022bridging}. These datasets align with previous work on UOS \cite{seitzer2022bridging, kakogeorgiou2024spot}, enabling a fair comparison. We refer to the supplementary material for further details. 
%on pre-processing and data splits.

\myparagraph{Metrics.} 
We employ standard metrics for unsupervised object segmentation. Following prior work \cite{kakogeorgiou2024spot}, we compute metrics on segmentation masks derived from both the slot-attention module and the decoder; we report the maximum across both in our experiments. The mean Intersection over Union ($\text{mIoU}$) quantifies segmentation accuracy by applying Hungarian matching between ground truth and predicted segmentation masks to maximize the IoU between segments on average. The mean Best Overlap ($\text{mBO}$) \cite{pont2016multiscale} assigns each ground-truth mask the predicted segment with the highest IoU, then averages over the assigned pairs. This metric has two variations: While $\text{mBO}^c$ requires ground-truth masks on the semantic, \ie class, level, $\text{mBO}^i$ is computed on instance, \ie object-level, ground-truth masks. Thus, $\text{mBO}^c$ is not applicable to the MOVi-C dataset, as the required annotations are not provided. We also report the Foreground Adjusted Rand Index (FG-ARI) to assess segmentation quality, but refer to concerns regarding its reliability \cite{kakogeorgiou2024spot, engelcke2020genesis}.

\myparagraph{Implementation.} For the multi-layer slot attention component, we choose the last four consecutive layers (see \cref{fig:hierarchy_with_SA}~(e)) of the encoder as input and fuse the extracted slot representations via M-Fusion. M-Fusion uses an MLP with one hidden layer of size 768 and GELU activations \cite{hendrycks2016bridging}. The number of slots depends on the dataset, where $K=6$ for VOC, $K=7$ for COCO, and $K=11$ for MOVi-C. The encoder utilizes a ViT-B/16 backbone, initialized with pre-trained DINO weights \cite{caron2021emerging}. Contrary to SPOT, we use segmentation masks of the SA module to distill knowledge of the teacher to the student, since we empirically found them to exhibit a higher segmentation quality (see supplement). Consistent with \cite{seitzer2022bridging}, we train for 1120 epochs on VOC, 100 epochs on COCO, and 95 epochs on MOVi-C. More details can be found in the supplementary material.

\subsection{Unsupervised object segmentation}
\begin{figure*}
    \centering
    \setlength{\tabcolsep}{1.5pt} % Adjust spacing between columns
    \begin{tabular}{cccccccccc} % First column for row labels, nine for images
       
        \raisebox{1.6\height}{\rotatebox[origin=c]{90}{\scriptsize \text{\ourspot}}} &
        \includegraphics[width=0.10350\linewidth]{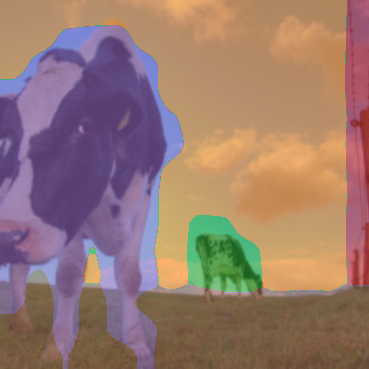} & 
        \includegraphics[width=0.10350\linewidth]{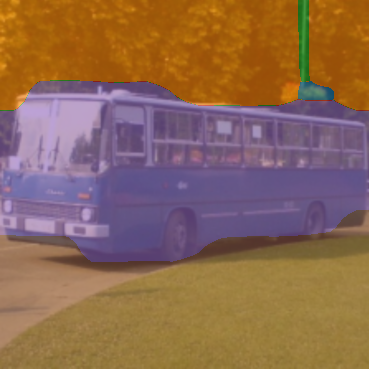} & 
        \includegraphics[width=0.10350\linewidth]{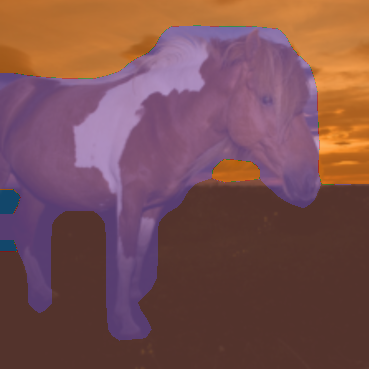} & 
        \includegraphics[width=0.10350\linewidth]{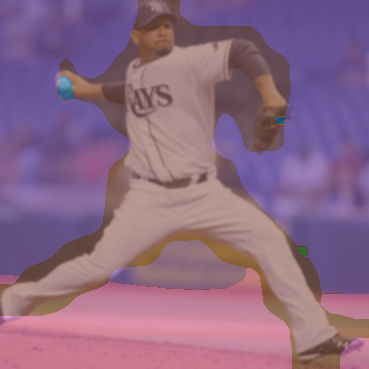} & 
        \includegraphics[width=0.10350\linewidth]{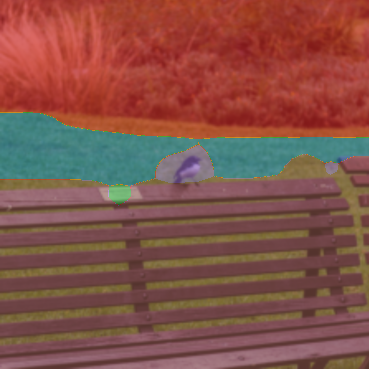} & 
        \includegraphics[width=0.10350\linewidth]{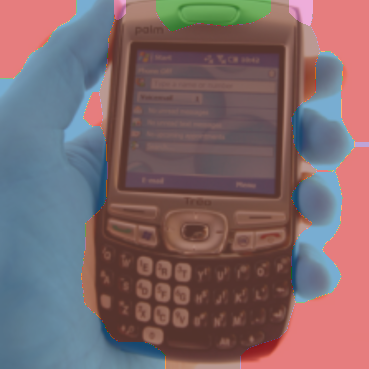} & 
        \includegraphics[width=0.10350\linewidth]{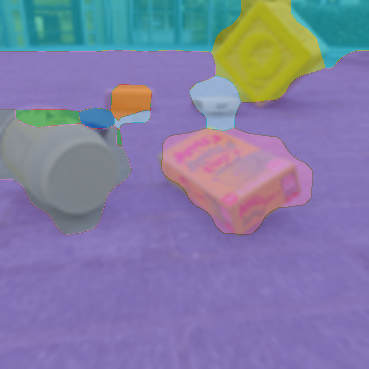} & 
        \includegraphics[width=0.10350\linewidth]{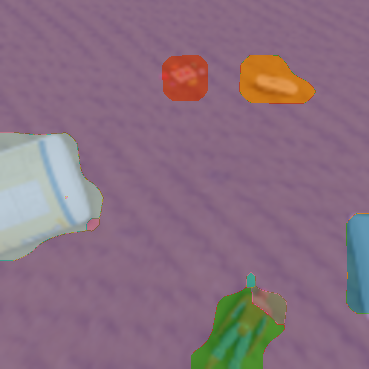} & 
        \includegraphics[width=0.10350\linewidth]{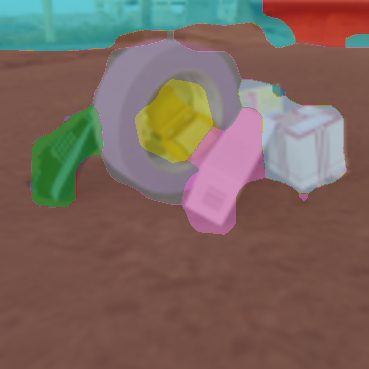} \vspace{-0.1em} \\
        
        % SPOT Row
        \raisebox{2.15\height}{\rotatebox[origin=c]{90}{\scriptsize \text{SPOT}}} &
        \includegraphics[width=0.1035\linewidth]{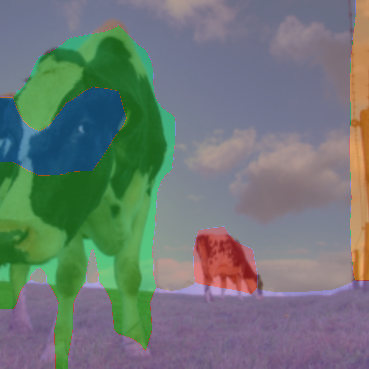} & 
        \includegraphics[width=0.1035\linewidth]{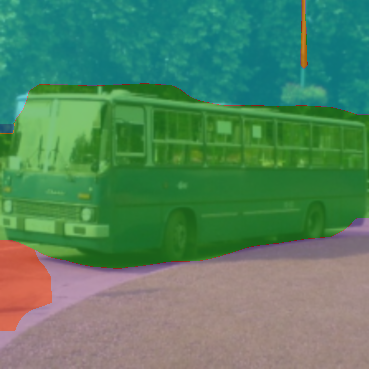} & 
        \includegraphics[width=0.1035\linewidth]{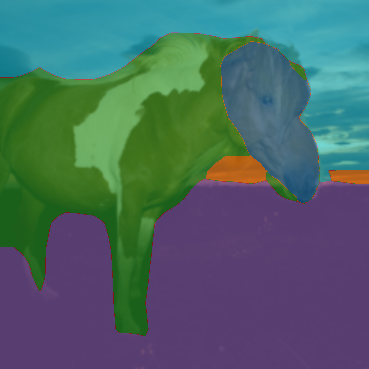} & 
        \includegraphics[width=0.1035\linewidth]{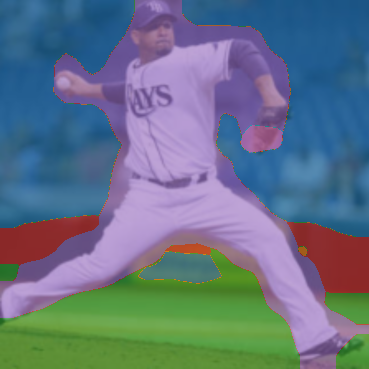} & 
        \includegraphics[width=0.1035\linewidth]{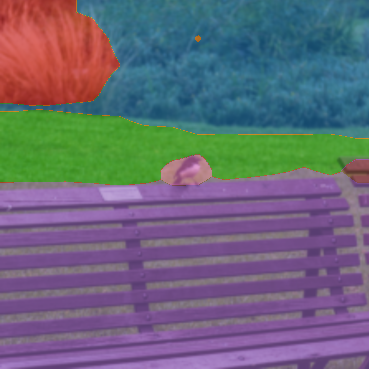} & 
        \includegraphics[width=0.1035\linewidth]{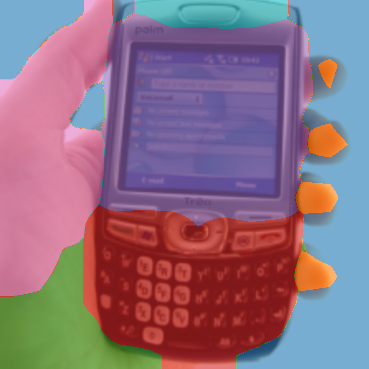} & 
        \includegraphics[width=0.1035\linewidth]{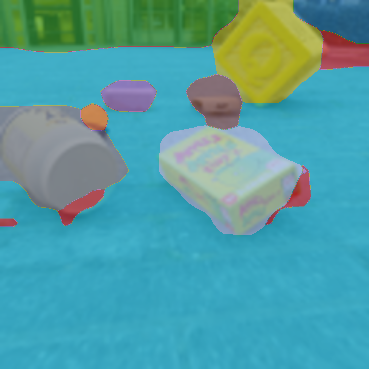} & 
        \includegraphics[width=0.1035\linewidth]{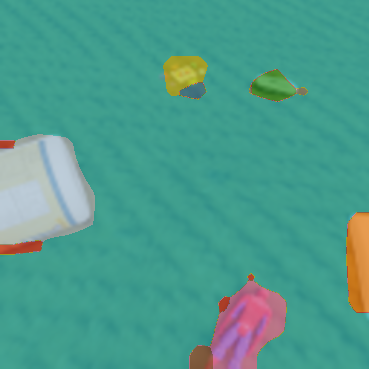} & 
        \includegraphics[width=0.1035\linewidth]{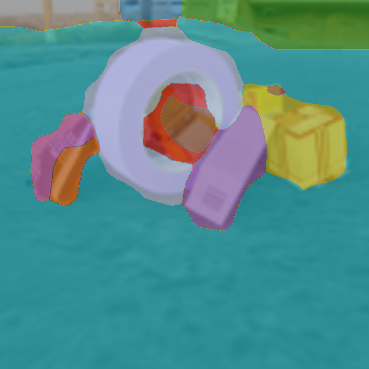} \vspace{-0.1em} \\

        % MUFASA-D Row
        \raisebox{0.935\height}{\rotatebox[origin=c]{90}{\scriptsize \text{\ourdino}}} &
        \includegraphics[width=0.1035\linewidth]    {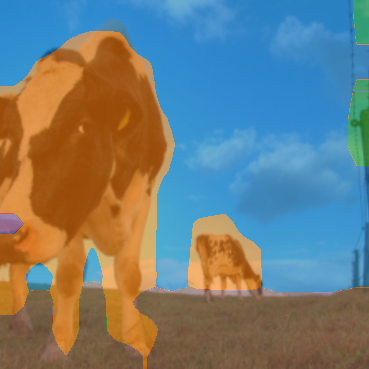} & 
        \includegraphics[width=0.1035\linewidth]{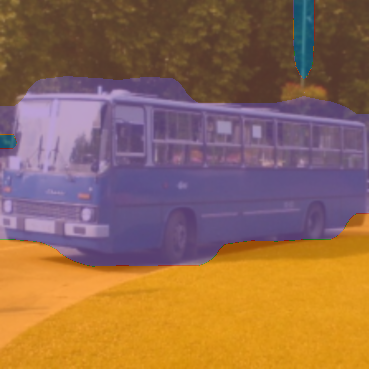} & 
        \includegraphics[width=0.1035\linewidth]{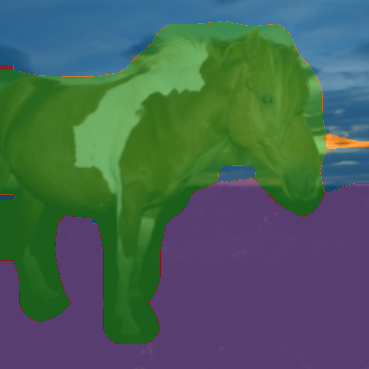} & 
        \includegraphics[width=0.1035\linewidth]{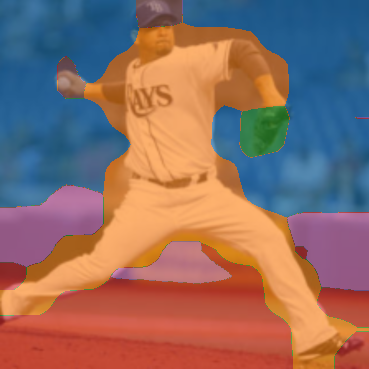} & 
        \includegraphics[width=0.1035\linewidth]{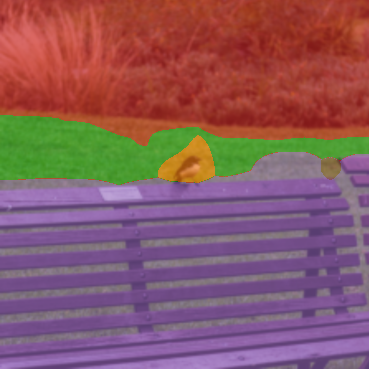} & 
        \includegraphics[width=0.1035\linewidth]{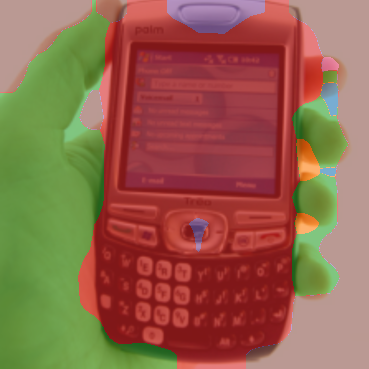} & 
        \includegraphics[width=0.1035\linewidth]{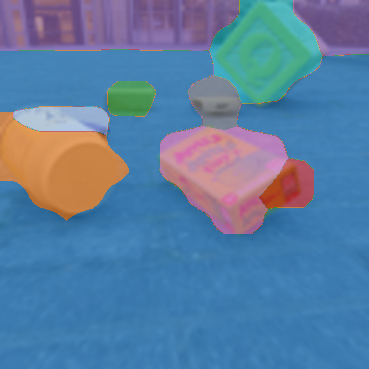} & 
        \includegraphics[width=0.1035\linewidth]{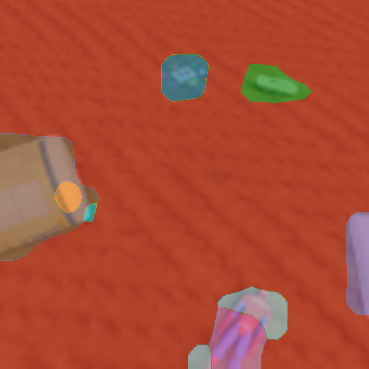} & 
        \includegraphics[width=0.1035\linewidth]{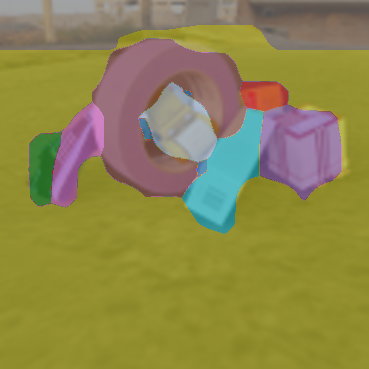} \vspace{-0.1em} \\
        
        % DINOSAUR Row
        \raisebox{1.15\height}{\rotatebox[origin=c]{90}{\scriptsize \text{DINOSAUR}}} &
       \includegraphics[width=0.1035\linewidth]{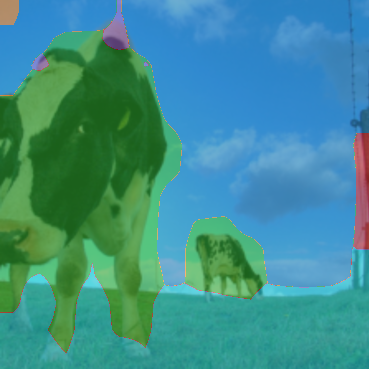} & 
        \includegraphics[width=0.1035\linewidth]{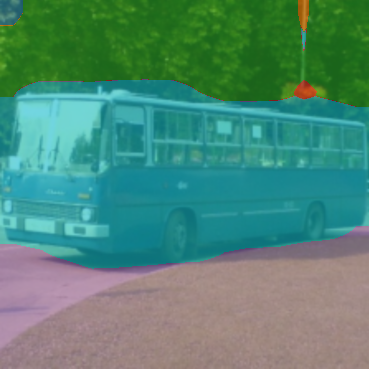} & 
        \includegraphics[width=0.1035\linewidth]{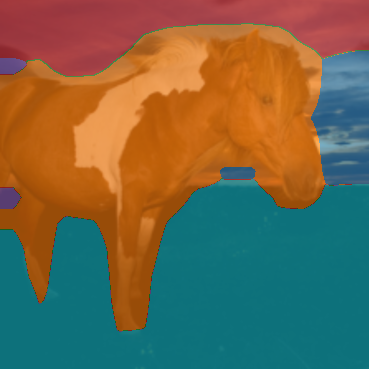} & 
        %frisbee, bench, 
        \includegraphics[width=0.1035\linewidth]{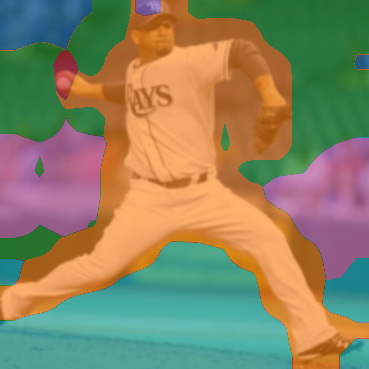} & 
        \includegraphics[width=0.1035\linewidth]{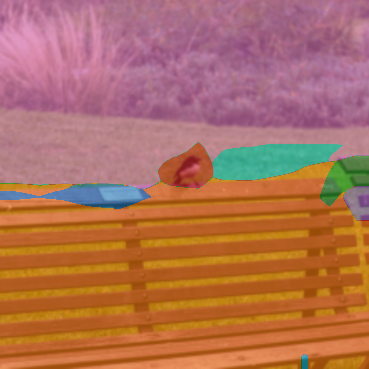} & 
        \includegraphics[width=0.1035\linewidth]{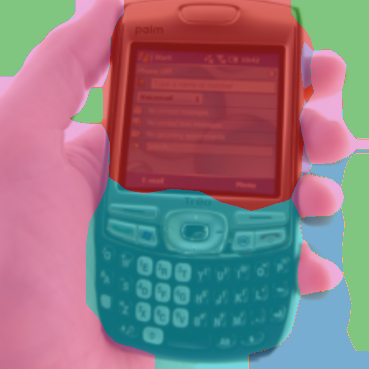} & 
        \includegraphics[width=0.1035\linewidth]{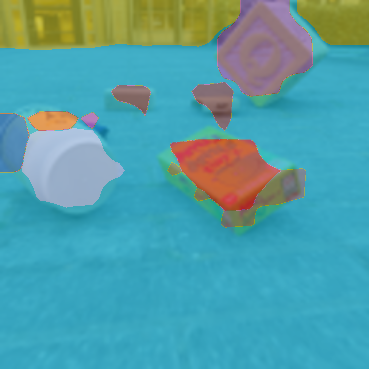} & 
        \includegraphics[width=0.1035\linewidth]{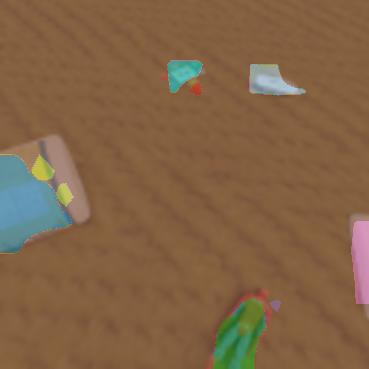} & 
        \includegraphics[width=0.1035\linewidth]{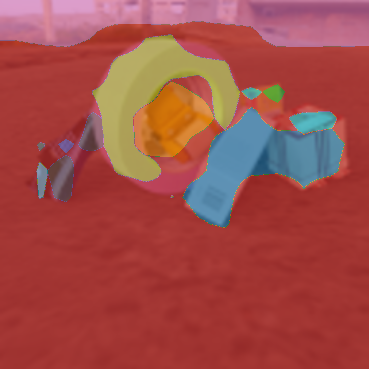} \vspace{-0.1em} \\
        
        % Ground Truth Row
        %\raisebox{0.79\height}{\rotatebox[origin=c]{90}{\small \text{Ground Truth}}} &
        %\includegraphics[width=0.1035\linewidth]{figures/ground_truths/cow.png} & 
        %\includegraphics[width=0.1035\linewidth]{figures/ground_truths/wheelie.png} & 
        %\includegraphics[width=0.1035\linewidth]{figures/ground_truths/dog.png} & 
        %\includegraphics[width=0.1035\linewidth]{figures/ground_truths/surfer.png} & 
        %\includegraphics[width=0.1035\linewidth]{figures/ground_truths/bench.png} & 
        %\includegraphics[width=0.1035\linewidth]{figures/ground_truths/phone.png} & 
        %\includegraphics[width=0.1035\linewidth]{figures/ground_truths/1_gt.png} & 
        %\includegraphics[width=0.1035\linewidth]{figures/ground_truths/2_gt.png} & 
        %\includegraphics[width=0.1035\linewidth]{figures/ground_truths/gt_movic_updated.png} \\

    \end{tabular}
    \vspace{-1em}
    \caption{\textbf{Comparison of segmentations.} Exemplary segmentation masks on nine different images for \ourspot \emph{(ours)}, SPOT, \ourdino \emph{(ours)}, and DINOSAUR. The first three images are from VOC, the next three from  COCO, and the last three from MOVi-C. Integrating \ourmethod results in segmentations that follow the object boundaries more closely than the baselines.}
    \label{fig:method_comparison}
    \vspace{-0.5em}
\end{figure*}
We integrate the \ourmethod framework into state-of-the-art OCL models \cite{seitzer2022bridging,kakogeorgiou2024spot} to highlight the benefit of the proposed multi-layer approach in UOS. As shown in \cref{tab:exp:mainresults_new}, both \ourspot and \ourdino consistently outperform their respective base models and prior methods across all datasets and metrics. With only one marginal exception in $\text{mIoU}$ on COCO, this establishes a comprehensive, new state of the art among unsupervised OCL methods on these benchmarks. 

The most substantial improvements across datasets are observed in class-level $\text{mBO}$ ($\text{mBO}^c$).
Notably, while \ourspot achieves the highest results overall on all datasets, even the less complex \ourdino surpasses the previous state of the art on PASCAL VOC and MOVi-C. This demonstrates the strength of the \ourmethod framework without depending on a sophisticated training setup as used in SPOT. On COCO, \ourdino improves upon DINOSAUR in all metrics, whereas \ourspot surpasses SPOT in $\text{mBO}^c$ and $\text{mBO}^i$  and ranks closely behind it in other metrics. 
Overall, these results indicate that leveraging multi-layer information enhances overall segmentation quality across most settings. Visually comparing the segmentation results of \ourspot and \ourdino against their base models (\cref{fig:method_comparison}), we see how our superior metrics correspond to visibly improved segmentations. In multiple cases, SPOT's and DINOSAUR’s segmentations miss object boundaries (\textit{col. 6}), split the object (\textit{col. 3}), or have holes in a segment (\textit{col. 1}). \ourspot and \ourdino produce masks consistent in shape and coverage, following the object boundaries more closely. 

\begin{figure}[t]
    \centering
    \resizebox{\linewidth}{!}{ % Stretch to full column width
    \begin{tabular}{@{}c@{\hskip 0.1in}c@{\hskip 0.1in}c@{\hskip 0.1in}c@{\hskip 0.1in}c@{}}
        \raisebox{0.32\height}{\huge $\hat{\mathcal{A}}^{\mathrm{Slot}}_1$} &  \raisebox{0.32\height}{\huge $\hat{\mathcal{A}}^{\mathrm{Slot}}_2$} &  \raisebox{0.32\height}{\huge $\hat{\mathcal{A}}^{\mathrm{Slot}}_3$} &  \raisebox{0.32\height}{\huge $\hat{\mathcal{A}}^{\mathrm{Slot}}_4$} & \raisebox{0.32\height}{\huge $\mathcal{A}^{\mathrm{Slot}}_{\mathrm{fused}}$} \\
        \includegraphics[width=0.6\linewidth,height=0.6\linewidth]{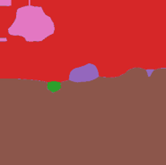} &
        \includegraphics[width=0.6\linewidth,height=0.6\linewidth]{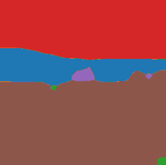} &
        \includegraphics[width=0.6\linewidth,height=0.6\linewidth]{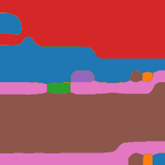} &
        \includegraphics[width=0.6\linewidth,height=0.6\linewidth]{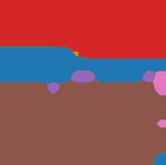} &
        \includegraphics[width=0.6\linewidth,height=0.6\linewidth]{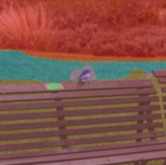}  \vspace{0.5em} \\
        \includegraphics[width=0.6\linewidth,height=0.6\linewidth]{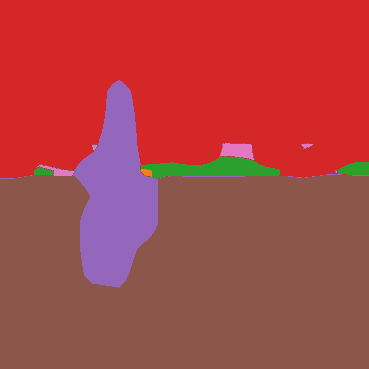} &
        \includegraphics[width=0.6\linewidth,height=0.6\linewidth]{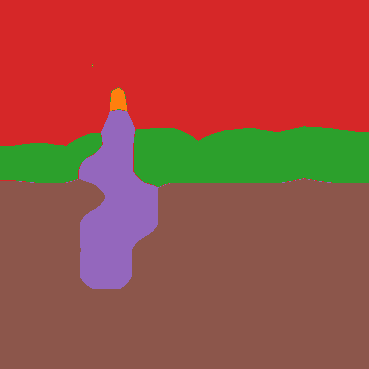} &
        \includegraphics[width=0.6\linewidth,height=0.6\linewidth]{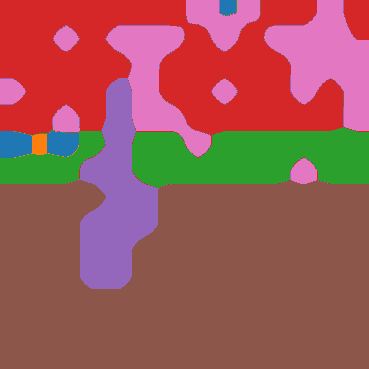} &
        \includegraphics[width=0.6\linewidth,height=0.6\linewidth]{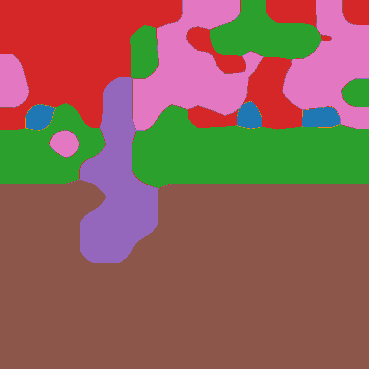} &
        \includegraphics[width=0.6\linewidth,height=0.6\linewidth]{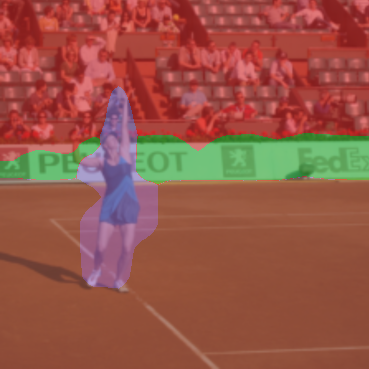} \\

    \end{tabular}
    }
    \vspace{-0.75em}
    \caption{\textbf{Segmentation per layer.} Layer-wise SA masks and the fused mask on COCO. Each layer contributes complementary information (\eg, \textit{row 1}: the plaque and bench edges in $\hat{\mathcal{A}}^{\mathrm{Slot}}_3$ \vs coarse segments in $\hat{\mathcal{A}}^{\mathrm{Slot}}_2$); the fused masks appear refined.}
    \label{fig:layerwise_ind}
    \vspace{-0.5em}
\end{figure}

In \cref{fig:layerwise_ind}, we present the attention masks from each slot-attention module alongside the final fused mask. Each layer contributes distinct information, and their combination produces a refined fused mask that more accurately delineates object boundaries than any individual layer mask. Notably, segmentation noise can appear at any layer and is not particularly confined to earlier or later layers. Nonetheless, the fused mask appears more accurate, suggesting that the complementary nature of layers compensates for each other's noise, resulting in a more complete segmentation.
\newcommand{\specialmidrule}{%
  \arrayrulecolor{gray}%
  \specialrule{0.3pt}{0pt}{0pt}
  %\hdashrule[0.5ex]{\linewidth}{0.4pt}{4pt 2pt}%
  \arrayrulecolor{black}%
}

\begin{table}[t]
    \renewcommand{\arraystretch}{1.15}  
    \caption{
    \textbf{Comparison of SA methods in UOS.} We compare our approach with OCL baselines on PASCAL VOC, COCO, and MOVi-C. The metrics (in \%, higher is better) are computed from slot-attention and decoder masks; the maximum across both is reported. For \ourmethod, we report mean $\pm$ std.\ dev.\ over three seeds. SA~\cite{locatello2020object} and SLATE~\cite{singh2021illiterate} results are taken from \cite{kakogeorgiou2024spot}. Results for SPOT \cite{kakogeorgiou2024spot} are reported without test-time ensembling (see supplement). Best results are in \textbf{bold}, the \nth{2}-best \underline{underlined}.
    }
    \label{tab:exp:mainresults_new}
    \vspace{-0.5em}
    \centering
    \scriptsize
    \setlength{\tabcolsep}{0.5pt} % Keep this in to align colored rows
    \begin{tabularx}{\columnwidth}{@{}
        >{\centering\arraybackslash}l
        >{\centering\arraybackslash}X
        >{\centering\arraybackslash}X
        >{\centering\arraybackslash}X
        >{\centering\arraybackslash}X
    @{}}  % Center-aligned columns
    \toprule
    Model & $\text{mBO}^c$ & $\text{mBO}^i$ & \text{mIoU} & \text{FG-ARI} \\
    \midrule
    \multicolumn{5}{c}{\raisebox{0.1em}{\textbf{PASCAL VOC}}} \\[-0.3em]  
    \midrule
    SA & 24.9 & 24.6 & \nodata & 12.3 \\
    SLATE & 41.5 & 35.9 & \nodata & 15.6 \\
    DINOSAUR & 51.2\spm{1.9} & 44.0\spm{1.9} & \nodata & \secondbest{24.8\spm{2.2}} \\
    \rowcolor{gray!15} \ourdino \ours & \secondbest{57.6\spm{0.6}} & \secondbest{49.2\spm{0.4}} & \secondbest{47.2\spm{0.5}}  & \best{25.2\spm{3.5}} \\
    SPOT & 55.3\spm{0.4} & 48.1\spm{0.4} & 46.5\spm{0.4} & 19.7\spm{0.4} \\
    \rowcolor{gray!15} \ourspot \ours & \best{59.8\spm{0.3}} & \best{51.3\spm{0.1}} & \best{49.4\spm{0.2}} & 20.6\spm{0.6} \\
    \midrule
    \multicolumn{5}{c}{\raisebox{0.3em}{\textbf{COCO}}}  \\ [-0.5em]  
    \midrule
    SA & 19.2 & 17.2 & \nodata & 21.4 \\
    SLATE & 33.6 & 29.1 & \nodata & 32.5 \\
    DINOSAUR & 39.7 & 31.6 & \nodata & \secondbest{34.1} \\
    \rowcolor{gray!15} \ourdino \ours & 43.0\spm{0.7} & 32.7\spm{0.4} & 30.5\spm{0.3} & 33.9\spm{1.1} \\
    SPOT & \secondbest{44.3\spm{0.3}} & \secondbest{34.7\spm{0.1}} & \best{32.7\spm{0.1}} & \best{37.8\spm{0.5}} \\
    \rowcolor{gray!15} \ourspot \ours & \best{45.5\spm{0.4}} & \best{34.8\spm{0.2}} & \secondbest{32.5\spm{0.2}} & 35.6\spm{0.9} \\
    \midrule
    \multicolumn{5}{c}{\raisebox{0.3em}{\textbf{MOVi-C}}}  \\[-0.5em]  
    \midrule
    SA & \nodata & 26.2\spm{1.0} & \nodata & 43.8\spm{0.3} \\
    SLATE & \nodata & 39.4\spm{0.8} & 37.8\spm{0.7} & 49.5\spm{1.4} \\
    DINOSAUR & \nodata & 42.4 & \nodata & 55.7 \\
    \rowcolor{gray!15} \ourdino \ours & \nodata & \best{49.2\spm{0.5}} & \best{48.3\spm{0.5}} & \secondbest{66.4\spm{2.1}}\\
    SPOT & \nodata & 47.0\spm{1.2} & 46.4\spm{1.2} & 57.9\spm{2.0} \\
    \rowcolor{gray!15} \ourspot \ours & \nodata & \best{49.2\spm{0.3}} & \secondbest{48.2\spm{0.3}} & \best{67.8\spm{1.7}} \\
    \bottomrule
    \end{tabularx}
    \vspace{-0.5em}
\end{table}

\subsection{Efficiency}
\label{sec:exp:efficiency}
Furthermore, we investigate the computational efficiency of our approach. \ourdino contains \SI{49.6}{\mega\nothing} trainable parameters, \SI{20.7}{\%} more than DINOSAUR (\SI{41.1}{\mega\nothing}) due to its multi-layer design. \ourspot consists of \SI{77.8}{\mega\nothing} trainable parameters, a relative increase of \SI{12.1}{\%} over SPOT (\SI{69.5}{\mega\nothing}). Therefore, we only modestly increase the number of trainable parameters. Nevertheless, our method comprehensively surpasses the results of baseline models, demonstrating its competitiveness in UOS. Even despite \ourdino having substantially fewer parameters than SPOT, it outperforms it on two out of three datasets. \ourmethod marginally increases the average GPU memory usage during training; with \ourdino requiring \SI{8.1}{\%} and  \ourspot \SI{0.4}{\%} more memory compared to their respective baselines. Additionally, we assess the training efficiency of our model in \cref{tab:exp:abl:efficiency}. We report results at two time instants, the number of epochs at which \emph{(i)} \ourspot{} / \ourdino first consistently surpass their baselines' results, and \emph{(ii)} when the peak results are achieved. Notably, our approach manages to perform on-par with baseline models in substantially fewer epochs, resulting in a \SI{94.4}{\%} reduction in training time for \ourspot and \SI{90.2}{\%} for \ourdino on VOC. Moreover, our models converge to a solution earlier on all datasets, suggesting that the multi-layer approach incorporates more information per image. Finally, we demonstrate inference efficiency by measuring throughput during evaluation. When integrating \ourmethod, we moderately reduce throughput from \SI{271.1}{\images\per\second} to \SI{217.3}{\images\per\second} for DINOSAUR and find a negligible reduction from \SI{86.1}{\images\per\second} to \SI{84.7}{\images\per\second} for SPOT. Consequently, \ourmethod is a lightweight addition to OCL, reaching SOTA results in UOS with reduced training times and minor computational overhead during inference, thus practically relevant for large-scale training.\begin{figure}[t]
    \centering
    \includegraphics[width=0.99\linewidth]{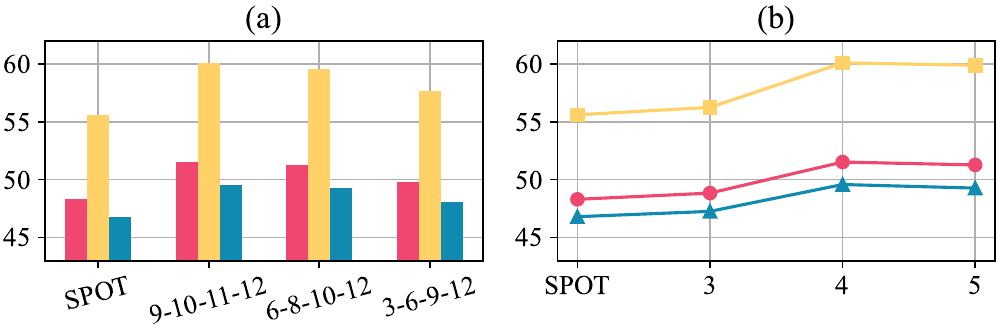}
    \vspace{-0.5em}
    \caption{\textbf{Ablations on layers}. \emph{(a)} Results of \ourspot on VOC in terms of \textcolor{pinklayer}{$\text{mBO}^i$}, \textcolor{yellowdinosaur}{$\text{mBO}^c$}, and \textcolor{bluespot}{$\text{mIoU}$} for different selections of layers compared to SPOT. \emph{(b)} Results of \ourspot on VOC in terms of \textcolor{pinklayer}{$\text{mBO}^i$}, \textcolor{yellowdinosaur}{$\text{mBO}^c$}, and \textcolor{bluespot}{$\text{mIoU}$ } for an increasing number of the last feature layers compared to the single-layer baseline SPOT \cite{kakogeorgiou2024spot}.}
    \label{fig:layerablation}
    \vspace{-0.5em}
\end{figure}\begin{table}
    \renewcommand{\arraystretch}{1.15} 
    \caption{\textbf{Training times}. We report the number of epochs to peak ($\text{E}_\text{Peak}$) and baseline-level ($\text{E}_\text{Base}$) results, defined as the first epoch exceeding baseline metrics ($\text{mBO}^c$ for VOC / COCO, $\text{mBO}^i$ for MOVi-C). The elapsed wall times for baseline  ($\text{T}_\text{Base}$) and peak results ($\text{T}_\text{Peak}$) are averaged over 100 epochs ($\text{T}_\text{Epoch}$). We outperform the baselines with fewer epochs and saturate faster \wrt wall time.}
    \label{tab:exp:abl:efficiency}
    \vspace{-0.5em}
    \centering
    \scriptsize
    \setlength{\tabcolsep}{0.5pt} % Keep this in to align colored rows
    \begin{tabularx}{\columnwidth}{@{}>{\centering\arraybackslash}l>{\centering\arraybackslash}X>{\centering\arraybackslash}X>
    {\centering\arraybackslash}X>
    {\centering\arraybackslash}X>
    {\centering\arraybackslash}X@{}}  % Center-aligned 
        \toprule
        Method  & $\text{E}_\text{Base}$ & $\text{E}_\text{Peak}$ & $\text{T}_\text{Epoch} (\text{min})$ & $\text{T}_\text{Base} (\text{h})$ & $\text{T}_\text{Peak} (\text{h})$\\
        \midrule
        \multicolumn{6}{c}{\raisebox{0.1em}{\textbf{PASCAL VOC}}} \\[-0.3em]  
        \midrule
         DINOSAUR & 644 & 644 & $\phantom{0}\text{2.0} \pm \text{0.4}$ & \phantom{0}21.5 & \phantom{0}21.5 \\
         \rowcolor{gray!15} \ourdino & \phantom{0}42 & 271 & $\phantom{0}\text{3.0} \pm \text{0.1}$ & \phantom{00}2.1 & \phantom{0}13.6 \\
         SPOT & 944 & 944 & $\phantom{0}\text{4.6} \pm \text{0.1}$  & \phantom{0}72.4 & \phantom{0}72.4 \\
         \rowcolor{gray!15} \ourspot & \phantom{0}51 & 615 & $\phantom{0}\text{4.8} \pm \text{0.1}$ & \phantom{00}4.1 & \phantom{0}49.2 \\
         \midrule
         \multicolumn{6}{c}{\raisebox{0.1em}{\textbf{COCO}}} \\[-0.3em]  
        \midrule
         DINOSAUR & \phantom{0}82 & \phantom{0}82 & $\text{80.7} \pm \text{2.3}$ & 110.3 & 110.3 \\
         \rowcolor{gray!15} \ourdino & \phantom{00}5 & \phantom{00}9 & $\text{81.2} \pm \text{5.8}$ & \phantom{00}6.8 & \phantom{0}12.2\\
         SPOT & \phantom{0}89 & \phantom{0}89 & $\text{84.6} \pm \text{9.0}$ & 125.5 & 125.5 \\
         \rowcolor{gray!15} \ourspot & \phantom{0}55 & \phantom{0}58 & $\text{84.9} \pm \text{9.2}$ & \phantom{0}77.8 & \phantom{0}82.1 \\
         \midrule
         \multicolumn{6}{c}{\raisebox{0.1em}{\textbf{MOVi-C}}} \\[-0.3em]  
        \midrule
         DINOSAUR & \phantom{0}63 & \phantom{0}63 & $\text{22.1} \pm \text{0.2}$ & \phantom{0}23.3 & \phantom{0}23.3 \\
         \rowcolor{gray!15} \ourdino & \phantom{00}7 & \phantom{0}29 & $\text{25.8} \pm \text{0.5}$ & \phantom{00}3.0 & \phantom{0}12.5 \\
         SPOT & \phantom{0}94 & \phantom{0}94 & $\text{32.9} \pm \text{0.2}$ & \phantom{0}51.5 & \phantom{0}51.5 \\
         \rowcolor{gray!15} \ourspot & \phantom{0}31 & \phantom{0}81 & $\text{34.5} \pm \text{0.3}$ & \phantom{0}17.8 & \phantom{0}46.6\\ 
         \bottomrule
    \end{tabularx}
    \vspace{-0.5em}
\end{table}

\subsection{Ablations}
\label{sec:ablations}
\paragraph{Layer choice ablation.} \label{sec:exp:layers} A key design choice for \ourmethod concerns the selection of feature layers on which to perform SA. Our findings, as shown in \cref{fig:hierarchy_with_SA}, indicate that earlier layers individually provide insufficient feature representations to be leveraged by slot attention. A PCA visualization, \cf supplement, further illustrates this semantic progression in the features.
This aligns with \cite{luo2024diffusionhyperfeaturessearchingtime}, which also found that the deepest DINO layers are best suited for semantic tasks. However, to examine whether earlier layers may provide effective complementary information, we investigate layer subsets that vary in terms of continuity, \ie consecutive \vs non-consecutive, combining different layer positions, including earlier ones.
Our ablation in \cref{fig:layerablation}~(a) shows that combining earlier with later layers indeed performs better than the baseline, but not better than using the last consecutive layers. 
Therefore, in \cref{fig:layerablation}~(b), we show the effect of varying the number of consecutive layers on \ourspot. We evaluate using the last three, four, and five feature layers. Results peak at four layers, and a further increase in the number of layers slightly deteriorates results, indicating that more layers do not necessarily translate to better segmentation.
Further considering the computational overhead each additional layer would entail, we achieve best segmentation results and efficiency through use of the last four consecutive layers for feature aggregation in \ourmethod. Yet, even for the other evaluated layer choices, \ourmethod consistently outperforms its respective baseline (\cf \cref{fig:layerablation}).

\myparagraph{Fusion strategy.} Within our framework, one main design decision concerns the fusion strategy, which integrates information from slots at multiple layers into a unified representation. In \cref{tab:exp:fusionstrategies}, we investigate alternative fusion strategies beyond M-Fusion. To this end, we first assess whether simply averaging of (matched) slot vectors and attention masks across layers \emph{(Avg-Fusion)} suffices. We find that Avg-Fusion performs worse compared to learned methods, indicating that non-learned fusion lacks the ability to aggregate multi-layer information sufficiently. We also examine a more complex learned strategy, where the MLP of M-Fusion is replaced with a transformer layer \emph{(T-Fusion)}. While T-Fusion surpasses SPOT, it does not outperform M-Fusion, suggesting that the additional complexity is not beneficial for slot fusion. Finally, we examine a variant of M-Fusion in which the pairwise summation of slot sets of adjacent layers prior to fusion is replaced with a plain concatenation \emph{(Concat-Fusion)}. Attention masks are analogously combined without this technique. We find that this leads to degraded results, demonstrating that the inductive bias of local interactions encoded by this pairwise summation effectively increases the performance of our fusion module.
\begin{table}[t]
    \renewcommand{\arraystretch}{1.15}         \caption{\textbf{Comparison of fusion strategies.} Segmentation metrics (in \%, $\uparrow$) of \ourspot on VOC with different fusion methods. Averaging matches SPOT, while M-Fusion performs best.
    }
    \label{tab:exp:fusionstrategies}
    \centering
    \scriptsize
    \setlength{\tabcolsep}{0.5pt} % Keep this in to align colored rows
    \vspace{-0.5em}
    \begin{tabularx}{\columnwidth}{@{}>{\raggedright\arraybackslash}p{2cm}>{\centering\arraybackslash}X>{\centering\arraybackslash}X>{\centering\arraybackslash}X>{\centering\arraybackslash}X@{}}  % Center-aligned columns
    \toprule
    Method & $\text{mBO}^c$ & $\text{mBO}^i$ & mIoU & FG-ARI \\
    \midrule
    SPOT & 55.3 & 48.1 & 46.5 & 19.7 \\
    Avg-Fusion & 55.6 & 48.1 & 46.5 & 19.4 \\
    Concat-Fusion & \secondbest{59.0} & \secondbest{50.9} & \secondbest{48.9} & \secondbest{20.0} \\
    T-Fusion & \secondbest{59.0} & 50.7 & \secondbest{48.9} & 19.7 \\
    \rowcolor{gray!15} M-Fusion \emph{(ours)} & \best{59.8} & \best{51.3} & \best{49.4} & \best{20.6} \\
    \bottomrule
    \end{tabularx}
    \vspace{-0.5em}
\end{table}

\myparagraph{Encoder choice.} Consistent with our baselines \cite{kakogeorgiou2024spot, seitzer2022bridging}, we use a ViT-B/16 as backbone, pre-trained with DINO \cite{caron2021emerging}, as the vision encoder to extract image features. Next, we evaluate \ourmethod with alternative pre-training schemes and encoder backbones, comprising MAE \cite{he2022masked} and DINOv2 \cite{oquab2023dinov2} features, as well as the ViT-S/8 and ViT-B/14 architectures. We restrict our study to these, since \cite{seitzer2022bridging} demonstrated that CNN-based encoders like ResNet yield substantially worse results in slot attention. Furthermore, the semantically rich representational hierarchy we leverage is innate to self-supervised ViTs \cite{amir2023on}. As shown in \cref{tab:exp:backbones}, \ourmethod consistently performs well across a variety of settings, outperforming its respective baselines in all instances, often by a substantial margin. These results demonstrate that the applicability of our approach is not confined to a specific encoder but generalizes across different architectures and pre-training schemes. Thus, \ourmethod is capable of utilizing different feature representations effectively within UOS.
\newcommand{\grayrow}[4]{%
    \cellcolor{gray!15}#1 & \cellcolor{gray!15}#2 & \cellcolor{gray!15}#3 & \cellcolor{gray!15}#4
}

\begin{table}[t]
    \renewcommand{\arraystretch}{1.15} 
    \caption{\textbf{Comparison of encoder backbones and pre-training schemes.} We report segmentation metrics (in \%, $\uparrow$) for \ourmethod and baselines utilizing different encoder backbones and pre-training schemes on VOC.
    (*) denotes values obtained from reproduction. We outperform the baselines across various encoder choices.
    }
    \label{tab:exp:backbones}
    \vspace{-0.5em}
    \centering
    \scriptsize
    \setlength{\tabcolsep}{0.5pt} % Keep this in to align colored rows
    \begin{tabularx}{\columnwidth}{@{}>{\centering\arraybackslash}l>{\centering\arraybackslash}l>{\centering\arraybackslash}l>{\centering\arraybackslash}X>{\centering\arraybackslash}X>{\centering\arraybackslash}X@{}}  % Center-aligned columns
    \toprule
    Weights \hspace{0.7em} & Backbone \hspace{0.7em} & Model & $\text{mBO}^c$ & $\text{mBO}^i$ & mIoU \\
    \midrule
    \multirow{4}{*}{DINO} & \multirow{4}{*}{ViT-B/16} & DINOSAUR & 51.2 & 44.0 & -- \\
                                             & & \grayrow{\ourdino}{\secondbest{57.6}}{\secondbest{49.2}}{\secondbest{47.2}} \\
                                                    & & SPOT     & 55.3 & 48.1 & 46.5 \\
                                             & & \grayrow{\ourspot}{\best{59.8}}{\best{51.3}}{\best{49.4}}  \\
    \midrule
    \multirow{4}{*}{MAE} & \multirow{4}{*}{ViT-B/16} & DINOSAUR$^*$ & 48.8 & 44.4 & 43.1 \\
                                             & & \grayrow{\ourdino}{\secondbest{55.0}}{46.8}{44.1} \\
                                                    & & SPOT$^*$    & 54.5 & \secondbest{47.4} & \secondbest{45.6} \\
                                             & & \grayrow{\ourspot}{\best{55.8}}{\best{48.8}}{\best{47.1}} \\
    \midrule
    \multirow{4}{*}{DINOv2} & \multirow{4}{*}{ViT-B/14} & DINOSAUR$^*$ & 49.5 & 43.1 & \secondbest{41.5} \\
                                             & & \grayrow{\ourdino}{\best{52.1}}{\best{44.4}}{\best{42.4}} \\
                                                    & & SPOT$^*$    & 49.4 & 43.0 & \secondbest{41.5} \\
                                             & & \grayrow{\ourspot}{\secondbest{51.0}}{\secondbest{44.2}}{\best{42.4}} \\
    \midrule
    \multirow{4}{*}{DINO} & \multirow{4}{*}{ViT-S/8} & DINOSAUR$^*$ & 51.2 & 45.9 & 44.7 \\
                                             & & \grayrow{\ourdino}{55.2}{\secondbest{48.8}}{\secondbest{47.5}} \\
                                                    & & SPOT$^*$    & \secondbest{55.3} & 48.4 & 47.0 \\
                                             & & \grayrow{\ourspot}{\best{60.0}}{\best{51.8}}{\best{49.7}} \\
    \bottomrule
    \end{tabularx}
    \vspace{-0.5em}
\end{table}

\myparagraph{Decoder choice.} In \cref{tab:exp:decoderimpact}, we explore the impact of an MLP decoder on the segmentation 
accuracy for DINOSAUR and SPOT, both with and without \ourmethod, noting that this decoder has weaker reconstruction abilities compared to the transformer decoder. We observe that the results of the baselines deteriorate more noticeably compared to \ourmethod. We attribute this robustness to the fact that our approach directly enhances the slot-attention mechanism, making it more agnostic to the choice of decoder. In contrast, SPOT increases its reliance on the decoder through its patch-order permutation strategy. Notably, \ourdino achieves better results than SPOT despite its simpler and more lightweight design, further demonstrating the strength of \ourmethod.
\begin{table}[t]
    \renewcommand{\arraystretch}{1.15} 
    \caption{
    \textbf{Decoder impact.} Results (in \%, $\uparrow$) from \ourmethod and baselines on COCO in UOS, using different decoder architectures. A weaker decoder (MLP) causes accuracy to deteriorate sharply; DINOSAUR and SPOT degrade more than our method.
    }
    \label{tab:exp:decoderimpact}
    \vspace{-0.5em}
    \centering
    \scriptsize
    \setlength{\tabcolsep}{0.5pt} % Keep this in to align colored rows
    \begin{tabularx}{\columnwidth}{@{}>{\centering\arraybackslash}l>{\centering\arraybackslash}X>{\centering\arraybackslash}X>{\centering\arraybackslash}X>{\centering\arraybackslash}X>{\centering\arraybackslash}X>{\centering\arraybackslash}X@{}}  % Center-aligned columns
    \toprule
    \multirow{2}{*}{Decoder} & \multicolumn{3}{c}{Transformer} & \multicolumn{3}{c}{MLP} \\  
    \cmidrule(lr){2-4} \cmidrule(lr){5-7}
    & $\text{mBO}^c$ & $\text{mBO}^i$ & mIoU & $\text{mBO}^c$ & $\text{mBO}^i$ & mIoU \\
    \midrule
    DINOSAUR & 39.7 & 31.6 & \nodata & 30.9 & 27.7 & \nodata \\
    \rowcolor{gray!15} \ourdino & 43.0 & 32.7 & 30.5 & \secondbest{34.0} & \secondbest{29.2} & \secondbest{27.9} \\
    SPOT & \secondbest{44.3} & \secondbest{34.7} & \best{32.7} & 32.4 & 28.4 & 27.0 \\
    \rowcolor{gray!15} \ourspot & \best{45.5} & \best{34.8} & \secondbest{32.5} & \best{34.7} & \best{30.2} & \best{28.9} \\
    \bottomrule
    \end{tabularx}
    \vspace{-0.5em}
\end{table}

\subsection{Limitations}
Inspecting class-level \vs instance-level segmentation masks, \ourmethod inherits some properties of previous slot-attention models as it tends to group instances of the same class into the same slot. This can be observed, \eg, when multiple people are shown in the scene (see supplement). This problem is not innate to \ourmethod but rather an issue of OCL models in general. In addition, Hungarian matching aligns slots across layers with a one-to-one mapping. Although effective, this may introduce a constraint, which makes flexible soft matching schemes a promising direction for future work.
% ideas? these are not really specific to our model
% 
\section{Conclusion}
\label{sec:conclusion}
We introduce \ourmethod, a multi-layer slot-attention mechanism that leverages the yet untapped information encoded within the feature layers of ViTs. As a lightweight, plug-and-play module, it can be easily integrated into existing slot-based methods for object-centric learning. \ourmethod also includes a novel method for effectively fusing slots from multiple layers into a unified representation. We demonstrate that incorporating \ourmethod leads to substantial gains in unsupervised object segmentation while also reducing training time. In this task, \ourspot sets a new state of the art on multiple benchmarks, highlighting its effectiveness and practical applicability.

\clearpage
\myparagraph{Acknowledgments.} This project has received funding from the European Research Council (ERC) under the European Union’s Horizon 2020 program (grant agreement No.\ 866008). The project was also supported by the Deutsche Forschungsgemeinschaft (DFG, German Research Foundation) under Germany's Excellence Strategy (EXC-3057/1 ``Reasonable Artificial Intelligence'', Project No.\ 533677015).
%in part by the State of Hesse through the “The Third Wave of Artificial Intelligence (3AI)” project.
Simone Schaub-Meyer has been funded by the DFG --~529680848.
Leonie Schüßler and Sebastian Bock are supported by the Konrad Zuse School of Excellence in Learning and Intelligent Systems \href{https://eliza.school/}{(ELIZA)} through the DAAD programme Konrad Zuse Schools of Excellence in Artificial Intelligence, sponsored by the German Federal Ministry of Education and Research.

{
    \small
    \bibliographystyle{ieeenat_fullname}
    \bibliography{main}
}

\clearpage
\pagestyle{plain}
\renewcommand{\thepage}{\roman{page}}
\setcounter{page}{1}
\maketitlesupplementary
%\setcounter{section}{0}
%\newbibstartnumber{70} % fill
%\setcounter{table}{6} % fill
%\setcounter{figure}{6} % fill 
%\renewcommand\thesection{\Alph{section}}
%\setcounter{page}{1}
%\pagenumbering{roman}
%\maketitlesupplementary
\appendix

%\twocolumn[%
%  \begin{@twocolumnfalse}
%    \centering {
%      \LARGE
%      \textbf{MUFASA: A Multi-Layer Framework for Slot Attention} \par
%      \vspace{0.75em}
%      \Large
%      Supplementary Material \par
%      \vspace{2.5em}
%    }
%  \end{@twocolumnfalse}
%]

% needs a limitation innate to \ourmethod? 

\section{Implementation Details}
In this section, we provide a more detailed overview of the training and implementation details for \ourdino and \ourspot. The relevant hyperparameters for every dataset are summarized in \cref{tab:appx:trn_dets}. In general, we train for \num{1120} epochs on VOC, 100 epochs on COCO, and 95 epochs on MOVi-C. While this remains consistent between \ourdino and \ourspot, the total number of epochs is split between the teacher and student model if self-training is employed. For the VOC and COCO datasets, the total number of epochs is evenly distributed between teacher and student, whereas for MOVi-C, the teacher is trained for 65 epochs following SPOT. We utilize the Adam optimizer \cite{kingmaAdam} with $\beta_0 = \num{0.9}$, $\beta_1 = \num{0.999}$ and no weight decay. Learning rate scheduling is employed using a linear warm-up to \num{10000} training steps and subsequent cosine annealing. For \ourspot on COCO, we empirically found the student to perform better with an increased warm-up for \num{30000} training steps. The learning rates are defined through a main value $\eta_\text{main}$ and a lower boundary $\eta_\text{low}$. In \ourspot, the learning rates of the students are set to match the teacher on VOC, whereas the peak value for COCO is reduced to $\eta_\text{main} = \num{3E-4}$ and the lower boundary for MOVi-C is set to $\eta_\text{low} = \num{1.5E-4}$. During self-training (\ie \ourspot), the knowledge distillation is incorporated into the reconstruction loss as the cross-entropy loss between aligned slot-attention masks of the teacher and student, weighed by some constant $\lambda$. We assign a greater weight to this loss as opposed to \cite{kakogeorgiou2024spot} with $\lambda = 0.01$, which we empirically found to work best for \ourspot. All experiments are conducted on a single NVIDIA RTX A6000 GPU with 48 GB of memory.

\subsection{MLP decoder}
In our ablations, we investigate the use of an MLP decoder for \ourmethod. Following previous work \cite{seitzer2022bridging, kakogeorgiou2024spot}, we implement it using a spatial broadcast decoder \cite{watters2019spatial}. Here, each of the $K$ slots in the fused slot representation is independently broadcast onto $N$ image patches. These patches correspond to the flattened $H_{\text{emb}} \times W_{\text{emb}}$ grid of the encoder, requiring the addition of learned positional encodings to convey the notion of order within it. Then, an MLP processes the image patches of each slot independently, converting them into meaningful feature information. This MLP is shared across all slots. In addition to the features that are constructed for every token, the MLP predicts unnormalized alpha values, determining how much a slot contributes to each image patch. This results in an independent feature reconstruction from each slot. To obtain attention masks for the MLP decoder, we normalize these alpha values across the slot dimension using softmax. Finally, the complete reconstruction is generated through a weighted linear combination of the slot features for every image patch, using the alpha masks as weights.
\begin{table}[t]
    \renewcommand{\arraystretch}{1.15} 
    \caption{
    \textbf{Slot-attention and decoder metrics using an MLP decoder.} UOS results (in \%, higher is better) of \ourmethod and baselines on COCO using an MLP decoder. ($\downarrow$) denotes the relative decrease in comparison to the transformer decoder. (*) indicates reproduced results. Decoder metrics degrade more substantially than slot-attention metrics when a weaker decoder is used.
    }
    \label{tab:exp:decoderslotmetrics}
    \vspace{-0.5em}
    \smallskip
    \centering
    \scriptsize
    \setlength{\tabcolsep}{0.5pt} % Keep this in to align colored rows
    \begin{tabularx}{\columnwidth}{@{}>{\centering\arraybackslash}l>{\centering\arraybackslash}X>{\centering\arraybackslash}X>{\centering\arraybackslash}X@{}}  % Center-aligned columns
    \toprule
    Model (MLP Dec.) & \hspace{2.2em}$\text{mBO}^c$ & \hspace{2.2em}$\text{mBO}^i$ & \hspace{1.8em}mIoU \\  
    \midrule
    \multicolumn{4}{c}{\raisebox{0.3em}{\textbf{Slot-Attention Metrics}}}  \\ [-0.5em]  
    \midrule
    DINOSAUR & 31.4$^*$\showdecrease{17.6} & 27.7$^*$\showdecrease{8.6} & 26.4$^*$ \\
    \rowcolor{gray!15} \ourdino \ours & 34.0\phantom{$^*$}\showdecrease{20.2} & 27.1\phantom{$^*$}\showdecrease{17.2} & 25.9\phantom{$^*$}\showdecrease{15.1} \\
    SPOT & 32.3$^*$\showdecrease{25.1} & 27.6$^*$\showdecrease{18.1} & 26.3$^*$\showdecrease{17.3} \\
    \rowcolor{gray!15} \ourspot \ours & 34.7\phantom{$^*$}\showdecrease{23.7} & 30.2\phantom{$^*$}\showdecrease{13.0} & 28.9\phantom{$^*$}\showdecrease{11.1} \\
    \midrule
    \multicolumn{4}{c}{\raisebox{0.3em}{\textbf{Decoder Metrics}}}  \\ [-0.5em]  
    \midrule
    DINOSAUR & 30.5$^*$\showdecrease{23.2} & 26.9$^*$\showdecrease{14.9} & 25.7$^*$ \\
    \rowcolor{gray!15} \ourdino \ours & 31.0\phantom{$^*$}\showdecrease{27.9} & 27.1\phantom{$^*$}\showdecrease{17.2} & 25.9\phantom{$^*$}\showdecrease{15.1} \\
    SPOT & 32.4\phantom{$^*$}\showdecrease{26.9} & 28.4\phantom{$^*$}\showdecrease{18.2} & 27.0\phantom{$^*$}\showdecrease{17.4}  \\
    \rowcolor{gray!15} \ourspot \ours & 32.0\phantom{$^*$}\showdecrease{25.4} & 27.5\phantom{$^*$}\showdecrease{18.2} & 26.3\phantom{$^*$}\showdecrease{16.9} \\
    \bottomrule
    \end{tabularx}
    \vspace{-0.7em}
\end{table}

\subsection{Visualization of segmentation masks}
To visualize segmentation masks (\eg, \cref{fig:appx_voc}), the fused attention mask $\mathcal{A}_{\mathrm{fused}}^{\mathrm{Slot}}$ is first reshaped to a spatial grid and upsampled to the image size with bilinear interpolation. Each pixel is assigned to the slot that attended most to it, where each slot is represented with a unique color. The resulting segmentation mask is then overlayed onto the image.
\subsection{Training stability for MAE and DINOv2}

Naively implementing MAE \cite{he2022masked} and DINOv2 \cite{oquab2023dinov2} as feature encoders leads to training collapse if no self-training is employed (\eg, in DINOSAUR and \ourdino). We mitigate this issue by using trainable initial slots instead of random initialization along with bi-level optimization (BOQSA \cite{BOQSA}). This strategy was originally introduced by \cite{kakogeorgiou2024spot} to stabilize training during image encoder fine-tuning.

\section{Discussion on Test-Time Ensembling}
\label{sec:appx:ensembling}
In their work, SPOT \cite{kakogeorgiou2024spot} employ test-time ensembling within the decoder by averaging predictions over nine decoder passes, one for each patch-order permutation. This yields  marginal improvements in their reported results at the cost of increased inference time. Given the minimal gains, we consider the additional inference cost unwarranted and therefore do not apply test-time ensembling. To ensure a fair comparison, we do not utilize test-time ensembling in either MUFASA or SPOT in our experiments.

\section{Discussion on Slot \textit{vs.}\ Decoder Masks}
\label{sec:appx:slotvsdecodermasks}

Empirically, integrating MUFASA yields stronger results when evaluating on segmentation masks derived from the slot attention module. In contrast, in previous models \cite{seitzer2022bridging, kakogeorgiou2024spot}, the decoder-produced masks were found more suitable for segmentation tasks. However, metrics computed based on decoder segmentations (\textit{decoder metrics}) are sensitive to the specific decoder architecture deployed. As shown in \cref{tab:exp:decoderslotmetrics}, when a weaker MLP decoder is used, the decoder metrics degrade substantially more than metrics computed on slot-attention segmentations (\textit{slot metrics}). This highlights the decoder capacity as a confounding factor. As a consequence, we observe that the decoder metrics do not reliably reflect the quality of the slot-object binding itself. By integrating MUFASA, we reduce this dependence and thereby reliably improve slot representations for UOS. Despite these limitations of decoder metrics, we report the maximum over both slot and decoder metrics in the main paper in accordance with prior work to enable a fair comparison.

\section{Comparison to Additional SOTA Models}
We provide further comparisons of \ourmethod against additional state-of-the-art models in the task of unsupervised object segmentation in \cref{tab:sota_comparison_uos}. Notably,~\cite{tian2025spotfsrc} relies on additional training signals beyond the reconstruction loss and \cite{slotadapt} leverages diffusion models pre-trained on caption-annotated data, while \ourmethod does not require any of these. Yet, our method outperforms them over multiple datasets and metrics.
\newcommand{\conflabel}[1]{\textcolor{gray}{{\fontsize{5.5}{6.5}\selectfont [#1]}}}

\begin{table}[t]
    \renewcommand{\arraystretch}{1.15} 
      \caption{
    \textbf{Comparison to additional SOTA methods in UOS.} We compare our approach with current SOTA OCL methods on PASCAL VOC, COCO, and MOVi-C. The metrics (in $\%$, higher is better) are computed from slot-attention and decoder masks; the maximum across both is reported. For \ourmethod, we report mean over three seeds. ``--'' indicates that results were not reported in the respective paper. We evaluate \cite{didolkar2025ftdinosaur} on the same resolution as \ourmethod. Best results are in \textbf{bold}, the \nth{2}-best \underline{underlined}.
    }
    \label{tab:sota_comparison_uos}
    \vspace{-0.7em}
    \centering
    \scriptsize
    \setlength{\tabcolsep}{0.5pt} % Keep this in to align colored rows
    \begin{tabularx}{\columnwidth}{@{}>{\centering\arraybackslash}l>{\centering\arraybackslash}X>{\centering\arraybackslash}X>{\centering\arraybackslash}X>{\centering\arraybackslash}X>{\centering\arraybackslash}X>{\centering\arraybackslash}X>{\centering\arraybackslash}X>{\centering\arraybackslash}X@{}}  % Center-aligned columns
    \toprule
    \multirow{2}{*}{Model} & \multicolumn{3}{c}{Pascal VOC} & \multicolumn{3}{c}{COCO} & \multicolumn{2}{c}{MOVi-C} \\  
    \cmidrule(lr){2-4} \cmidrule(lr){5-7} \cmidrule(lr){8-9}
    & $\text{mBO}^c$ & $\text{mBO}^i$ & mIoU & $\text{mBO}^c$ & $\text{mBO}^i$ & mIoU & $\text{mBO}^i$ & mIoU \\
    \midrule
    SlotAdapt \cite{slotadapt}  & 51.9 & \best{51.5} & \nodata & 39.2 & \secondbest{35.1} & \nodata & \nodata & \nodata \\
    Multi-Query SA \cite{pramanik2024masked} & \nodata & 39.7 & 39.4 & \nodata & \nodata & \nodata & \nodata & \nodata \\
    FT-DINOSAUR \cite{didolkar2025ftdinosaur}  & \nodata & \nodata & \nodata & \nodata & 32.0 & \nodata & \nodata & \nodata \\
    SPOT-FS-RC \cite{tian2025spotfsrc}  & 56.5 & 49.3 & \nodata & \secondbest{45.3} & \best{35.7} & \nodata & 49.0 & 47.8 \\
    \rowcolor{gray!15} \ourdino \emph{(ours)} & \secondbest{57.6} & 49.2 & \secondbest{47.2} & 43.0 & 32.7 & \secondbest{30.5} & \best{49.2} & \best{48.3} \\
    \rowcolor{gray!15} \ourspot \emph{(ours)} & \best{59.8} & \secondbest{51.3} & \best{49.4} & \best{45.5} & 34.8 & \best{32.5} & \best{49.2} & \secondbest{48.2} \\
    \bottomrule
    \end{tabularx}
    \vspace{-0.75em}
  
\end{table}

\newcolumntype{R}[1]{>{\raggedleft\arraybackslash}p{#1}}
\begin{table*}[h]
\captionsetup{width=\linewidth}  
\caption{\textbf{Hyperparameters of MUFASA on the VOC, COCO, and MOVi-C datasets.} Learning rates and warmup epochs may differ between teacher and student if self-training is employed; students utilizing a different learning rate and warmup schedule than their teacher are denoted with $\dagger$. ``--/--'' denotes identical hyperparameters across datasets.}
\label{tab:appx:trn_dets}
\scriptsize
\setlength{\tabcolsep}{2pt} % 
\centering

\begin{tabularx}{\textwidth}{XX R{3.75cm} R{3.75cm} R{3.75cm}}
\toprule
\textbf{Dataset} $\rightarrow$ &  & \textbf{PASCAL VOC} &  \textbf{COCO} & \textbf{MOVI-C} \\
\midrule
\multirow{4}{*}{Epochs} & Teacher & 560 & 50 & 65 \\
& Student & 560 & 50 & 30 \\
& No self-training & \num{1120} & 100 & 95\\
& Warmup & 60 & 5 ($\dagger:$ 15) & 7 \\
\midrule
Low LR $\eta_{\text{low}}$ & & \num{4E-7} & \num{4E-7} & \num{4E-5} ($\dagger$: \num{1.5E-4}) \\
Main LR $\eta_{\text{main}}$ & & \num{4E-4} & \num{4E-4} ($\dagger$: \num{3E-4}) & \num{2E-4} \\
Batch size & & 64 & --/--  & --/--  \\
Optimizer & & Adam ($\beta_0=\num{0.9}$, $\beta_1=\num{0.999}$) & --/-- & --/--  \\
Distillation $\lambda$ & & 0.01 & --/-- & --/-- \\
\midrule
\multirow{4}{*}{Encoder} & Architecture & ViT-B \cite{visiontransformer} & --/-- & --/-- \\
 & Patch size & 16 $\times$ 16 & --/-- & --/-- \\
 & Feature dimension $d_{\text{emb}}$ & 768 & --/-- & --/-- \\
 & Weights & DINO \cite{caron2021emerging} & --/-- & --/-- \\
\midrule
\multirow{2}{*}{ViT decoder} & Number of layers & 4 & --/-- & --/-- \\
 & Heads & 6 & --/-- & --/-- \\
 \midrule
\multirow{2}{*}{MLP decoder} & Number of layers & 4 & --/-- & --/--  \\
 & Hidden size & \num{2048} & --/-- & --/--  \\
\midrule
\multirow{4}{*}{Slot fusion} & Strategy & M-Fusion & --/-- & --/-- \\
 & Layer selection $\mathcal{J}$ & 9, 10, 11, 12 & --/-- & --/-- \\
 & MLP hidden layers & 1 & --/-- & --/-- \\
 & Activation & GELU \cite{hendrycks2016bridging} & --/-- & --/-- \\
 & MLP hidden size & 768 & --/-- & --/-- \\
\midrule
\multirow{4}{*}{Slot attention} & Iterations & 3 & --/-- & --/-- \\
 & MLP hidden size & \num{1024} & --/-- & --/--\\
 & Slot dimension $d_{\text{slot}}$ & 256 & --/-- & --/-- \\
 & Number of slots & 6 & 7 & 11 \\
\midrule
 Training images & & \num{10582} & \num{118287} & \num{87633}  \\
 Evaluation images & & \num{1449} & \num{5000} & \num{6000} \\
 Crop resolution & & 224 $\times$ 224 & --/-- & --/-- \\
 Evaluation resolution & & 320 $\times$ 320 & 320 $\times$ 320 & 128 $\times$ 128 \\
Resize strategy & & Minor axis to 224 & Minor axis to 224 & -- \\
Crop strategy & & Random & Center & Full \\
Augmentations & & Random flip ($p=\num{0.5}$) & Random flip ($p=\num{0.5}$) & -- \\

\bottomrule
\end{tabularx}
\vspace{-0.5em}

\end{table*}

\section{Visualization of ViT Layers}
\begin{figure*}
    \centering
    \setlength{\tabcolsep}{2pt} 
    \renewcommand{\arraystretch}{0.3} 
    \begin{tabular}{ccccccc}
        % First Row
        \begin{minipage}{0.13\linewidth} \centering
            \includegraphics[width=\linewidth]{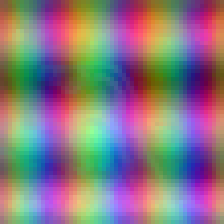}
        \end{minipage} & 
        \begin{minipage}{0.13\linewidth} \centering
            \includegraphics[width=\linewidth]{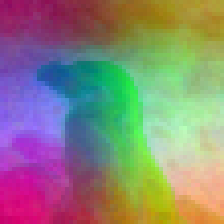}
        \end{minipage} & 
        \begin{minipage}{0.13\linewidth} \centering
            \includegraphics[width=\linewidth]{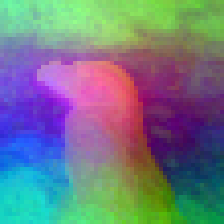}
        \end{minipage} & 
        \begin{minipage}{0.13\linewidth} \centering
            \includegraphics[width=\linewidth]{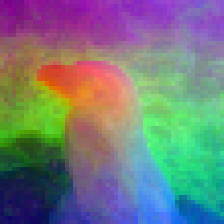}
        \end{minipage} & 
        \begin{minipage}{0.13\linewidth} \centering
            \includegraphics[width=\linewidth]{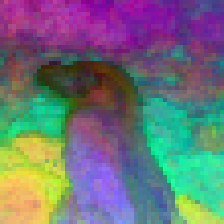}
        \end{minipage} & 
        \begin{minipage}{0.13\linewidth} \centering
            \includegraphics[width=\linewidth]{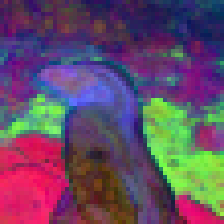}
        \end{minipage} & 
        \begin{minipage}{0.13\linewidth} \centering
            \includegraphics[width=\linewidth]{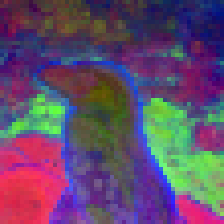}
        \end{minipage}\\ \\
        
        % Second Row
        \begin{minipage}{0.13\linewidth} \centering
            \includegraphics[width=\linewidth]{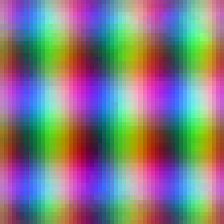}
        \end{minipage} & 
        \begin{minipage}{0.13\linewidth} \centering
            \includegraphics[width=\linewidth]{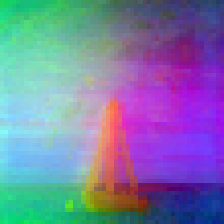}
        \end{minipage} & 
        \begin{minipage}{0.13\linewidth} \centering
            \includegraphics[width=\linewidth]{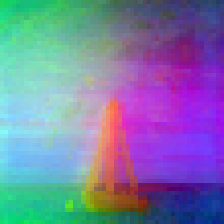}
        \end{minipage} & 
        \begin{minipage}{0.13\linewidth} \centering
            \includegraphics[width=\linewidth]{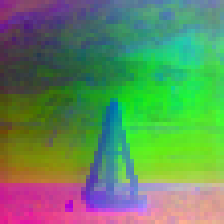}
        \end{minipage} & 
        \begin{minipage}{0.13\linewidth} \centering
            \includegraphics[width=\linewidth]{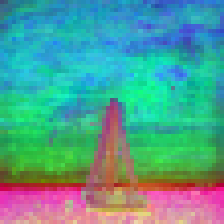}
        \end{minipage} &  
        \begin{minipage}{0.13\linewidth} \centering
            \includegraphics[width=\linewidth]{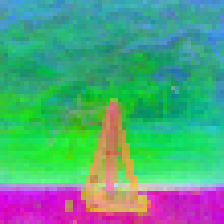}
        \end{minipage} &  
        \begin{minipage}{0.13\linewidth} \centering
            \includegraphics[width=\linewidth]{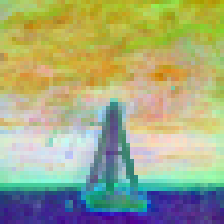}
        \end{minipage} \\ \\

        % Third Row
        \begin{minipage}{0.13\linewidth} \centering
            \includegraphics[width=\linewidth]{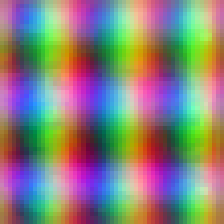}
        \end{minipage} &  
        \begin{minipage}{0.13\linewidth} \centering
            \includegraphics[width=\linewidth]{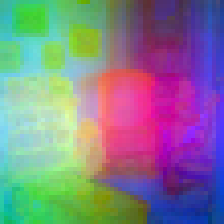}
        \end{minipage} &  
        \begin{minipage}{0.13\linewidth} \centering
            \includegraphics[width=\linewidth]{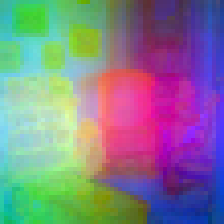}
        \end{minipage} &  
        \begin{minipage}{0.13\linewidth} \centering
            \includegraphics[width=\linewidth]{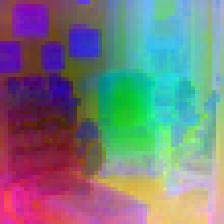}
        \end{minipage} &  
        \begin{minipage}{0.13\linewidth} \centering
            \includegraphics[width=\linewidth]{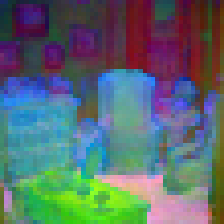}
        \end{minipage} &  
        \begin{minipage}{0.13\linewidth} \centering
            \includegraphics[width=\linewidth]{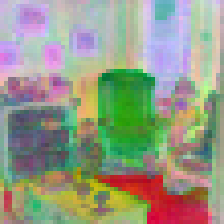}
        \end{minipage} &  
        \begin{minipage}{0.13\linewidth} \centering
            \includegraphics[width=\linewidth]{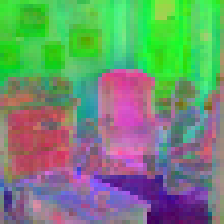}
        \end{minipage} \\ \\

        % Fourth Row (Last Row with Subcaptions)
        \begin{minipage}{0.13\linewidth} \centering
            \includegraphics[width=\linewidth]{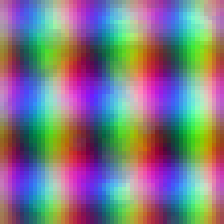}
            \subcaption{Layer 3}
        \end{minipage} &  
        \begin{minipage}{0.13\linewidth} \centering
            \includegraphics[width=\linewidth]{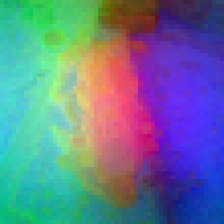}
            \subcaption{Layer 6}
        \end{minipage} &  
        \begin{minipage}{0.13\linewidth} \centering
            \includegraphics[width=\linewidth]{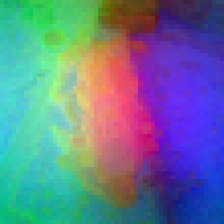}
            \subcaption{Layer 8}
        \end{minipage} &  
        \begin{minipage}{0.13\linewidth} \centering
            \includegraphics[width=\linewidth]{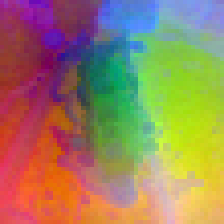}
            \subcaption{Layer 9}
        \end{minipage} &  
        \begin{minipage}{0.13\linewidth} \centering
            \includegraphics[width=\linewidth]{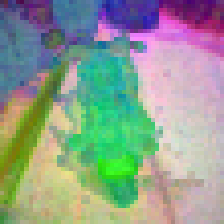}
            \subcaption{Layer 10}
        \end{minipage} &  
        \begin{minipage}{0.13\linewidth} \centering
            \includegraphics[width=\linewidth]{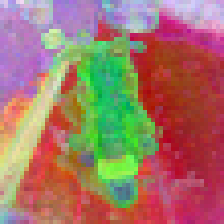}
            \subcaption{Layer 11}
        \end{minipage} &  
        \begin{minipage}{0.13\linewidth} \centering
            \includegraphics[width=\linewidth]{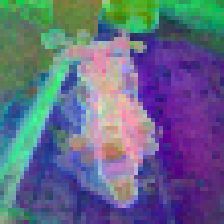}
            \subcaption{Layer 12}
        \end{minipage}  
    \end{tabular}
    \caption{\textbf{PCA of DINO ViT features}. Layerwise visualization of the DINO ViT features at different layers via principal component analysis (PCA) for four different images. The first three principal components yield red, green, and blue channels. Semantically meaningful information is absent in earlier layers and begins to emerge in intermediate ones, while becoming increasingly rich in deeper layers. }
    \label{fig:appendix_vit_hierarchy}
\end{figure*}

The feature representations at different layers of the DINO ViT \cite{caron2020unsupervised} serve as the foundation for \ourmethod's multi-layer slot attention. In this section, we analyze how their structural properties and encoding characteristics evolve across layers. To do so, we conduct a principal component analysis (PCA), visualized for all layers that were investigated in our ablations on layer choice. 
Following \cite{amir2023on}, we project the high-dimensional feature representations to three principal components, which are then mapped to RGB channels for visualization.
In \cref{fig:appendix_vit_hierarchy} (a), we observe a grid-like structure at layer 3, devoid of any object-specific shape. This suggests that such early layers are unsuitable as input to slot attention, as they lack object-centric information, visually confirming our ablations in \cref{fig:layerablation}. In the intermediate layers (\cref{fig:appendix_vit_hierarchy} (b) -- (d)), the object structure gradually emerges, and background textures become distinguishable. These layers provide information about the object localization. At last, the latest layers (\cref{fig:appendix_vit_hierarchy} (e) -- (g)) exhibit semantic information, such as the stripe in the fur at the penguin's head (first row) or the small items on the table (third row). At this stage, the characteristics are now semantically meaningful features to form object-centric representations. 

\section{Additional Visual Examples}
\begin{figure*}
    \centering
    \setlength{\tabcolsep}{2pt} % Adjust spacing between columns
    \begin{tabular}{ccccccccc} % First column for row labels, eight for images
        % MUFASA ROW
        \raisebox{1.7\height}{\rotatebox[origin=c]{90}{\scriptsize \text{\ourspot}}} &
        \includegraphics[width=0.11\linewidth]{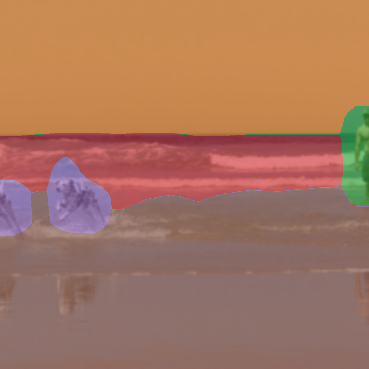} &
        \includegraphics[width=0.11\linewidth]{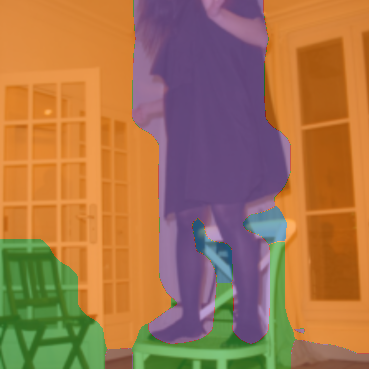} &
        \includegraphics[width=0.11\linewidth]{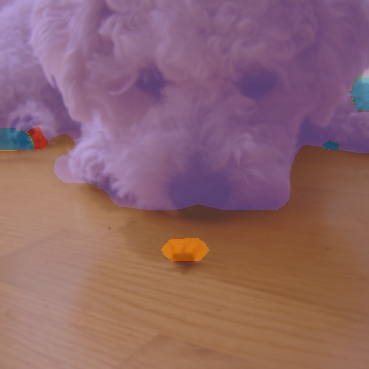} &
        \includegraphics[width=0.11\linewidth]{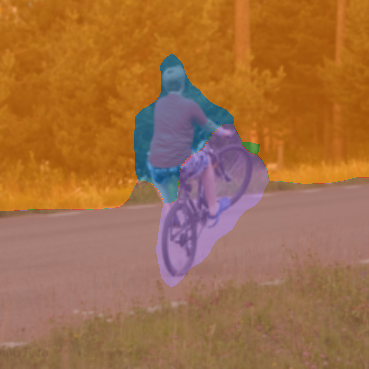} &
        \includegraphics[width=0.11\linewidth]{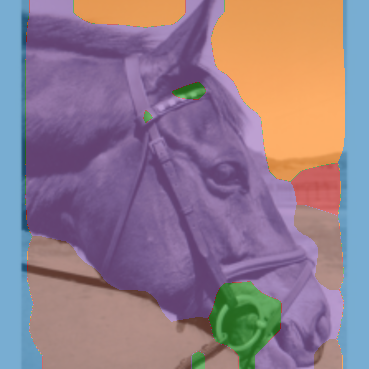} &
        \includegraphics[width=0.11\linewidth]{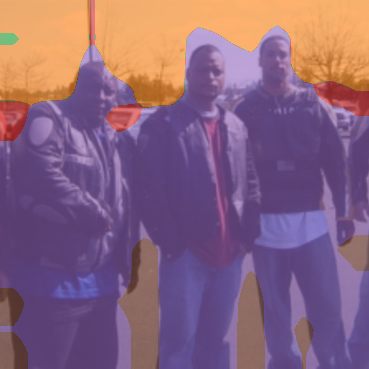} &
        \includegraphics[width=0.11\linewidth]{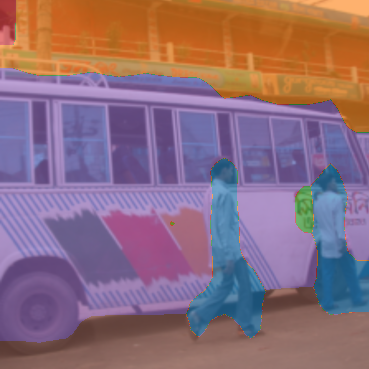} &
        \includegraphics[width=0.11\linewidth]{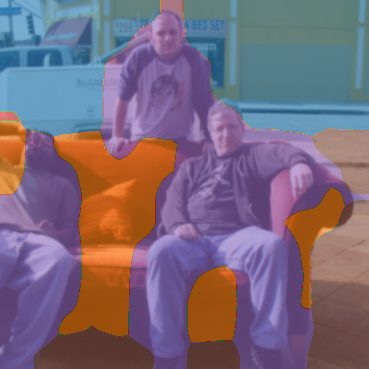} \\
        
        % SPOT ROW
        \raisebox{2.35\height}{\rotatebox[origin=c]{90}{\scriptsize \text{SPOT}}} &
        \includegraphics[width=0.11\linewidth]{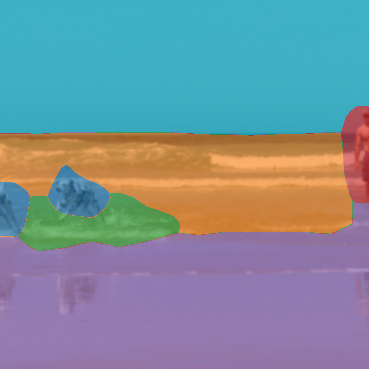} &
        \includegraphics[width=0.11\linewidth]{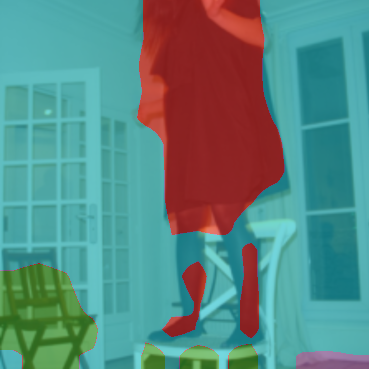} &
        \includegraphics[width=0.11\linewidth]{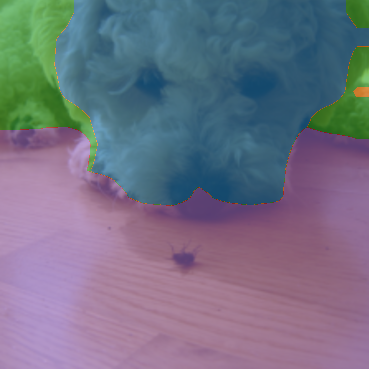} &
        \includegraphics[width=0.11\linewidth]{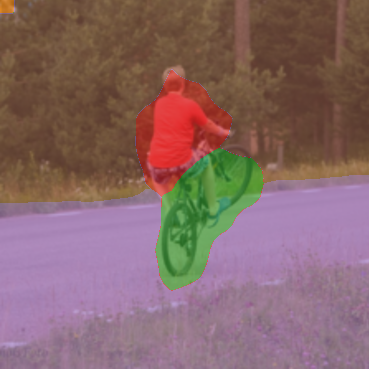} &
        \includegraphics[width=0.11\linewidth]{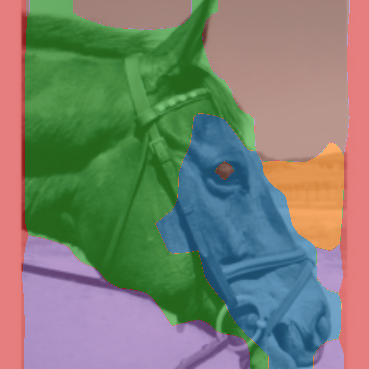} &
        \includegraphics[width=0.11\linewidth]{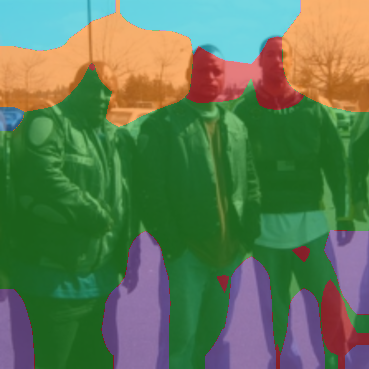} &
        \includegraphics[width=0.11\linewidth]{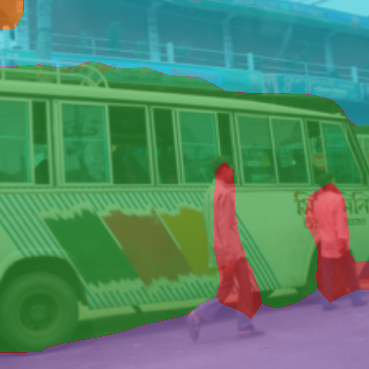} &
        \includegraphics[width=0.11\linewidth]{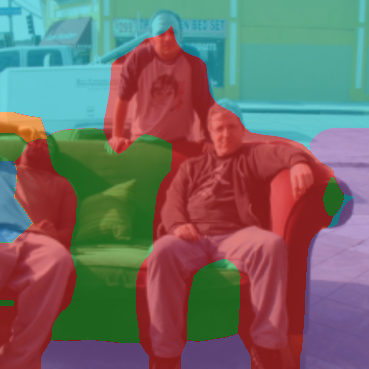} \\

        % MUFASA-D ROW
        \raisebox{1.05\height}{\rotatebox[origin=c]{90}{\scriptsize \text{\ourdino}}} &
        \includegraphics[width=0.11\linewidth]{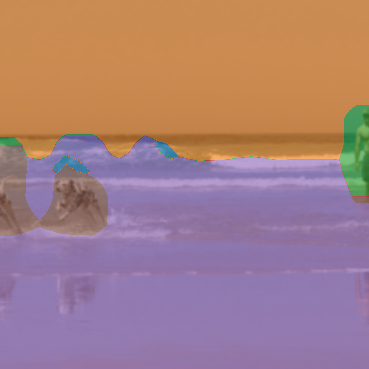} &
        \includegraphics[width=0.11\linewidth]{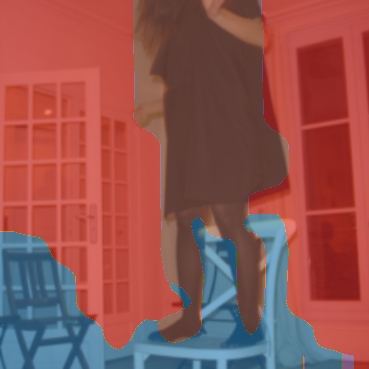} &
        \includegraphics[width=0.11\linewidth]{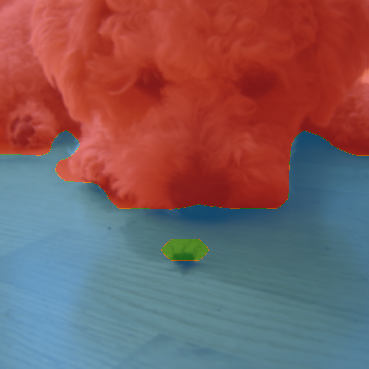} &
        \includegraphics[width=0.11\linewidth]{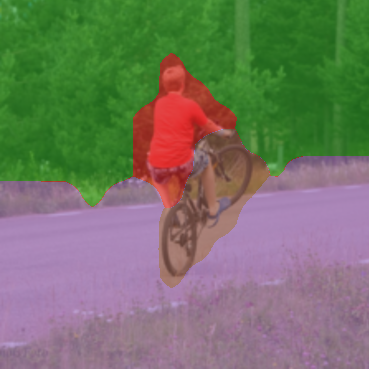} &
        \includegraphics[width=0.11\linewidth]{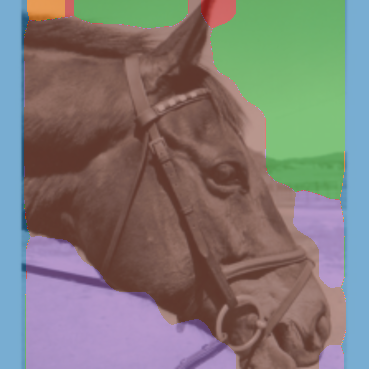} &
        \includegraphics[width=0.11\linewidth]{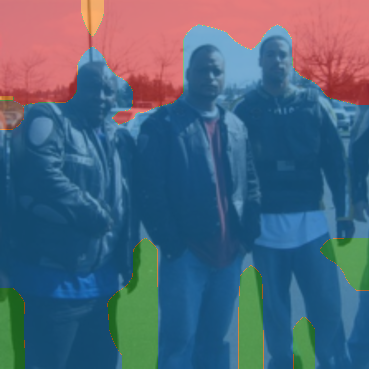} &
        \includegraphics[width=0.11\linewidth]{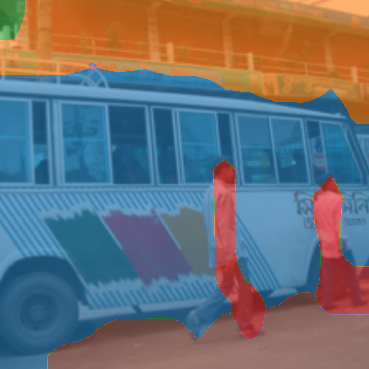} &
        \includegraphics[width=0.11\linewidth]{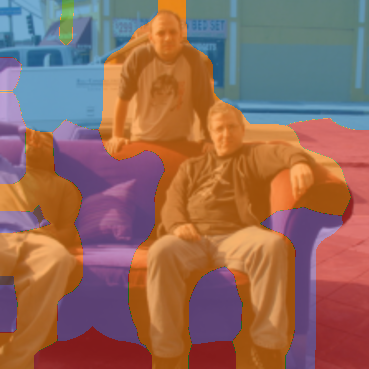} \\
        
        % DINOSUR ROW
        \raisebox{1.25\height}{\rotatebox[origin=c]{90}{\scriptsize \text{DINOSAUR}}} &
        \includegraphics[width=0.11\linewidth]{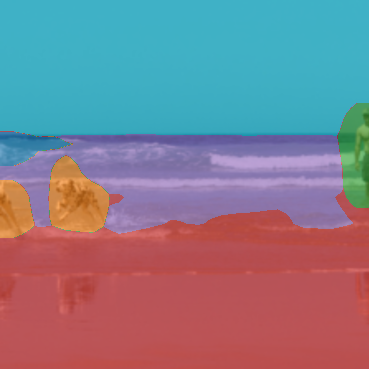} &
        \includegraphics[width=0.11\linewidth]{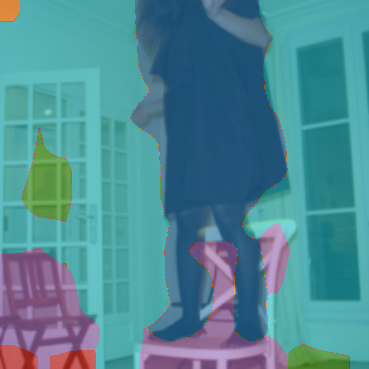} &
        \includegraphics[width=0.11\linewidth]{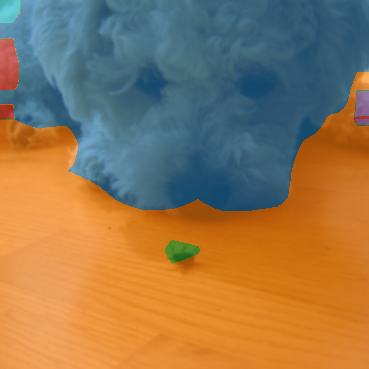} &
        \includegraphics[width=0.11\linewidth]{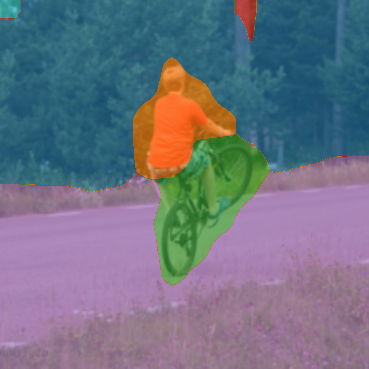} &
        \includegraphics[width=0.11\linewidth]{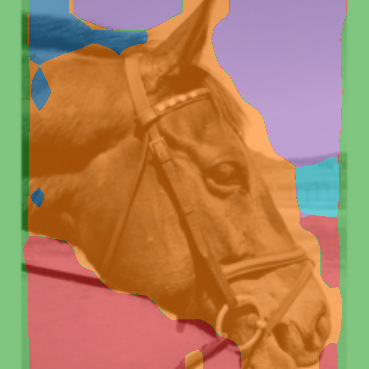} &
        \includegraphics[width=0.11\linewidth]{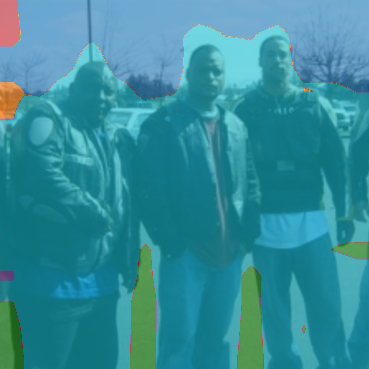} &
        \includegraphics[width=0.11\linewidth]{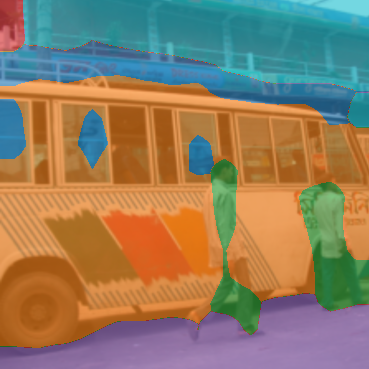} &
        \includegraphics[width=0.11\linewidth]{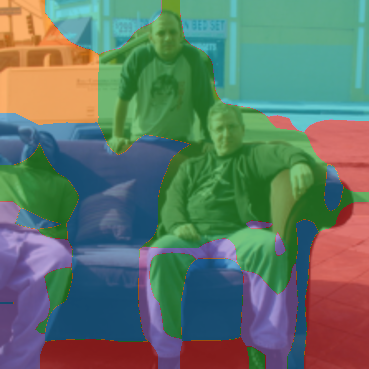} \\
        
        % GT ROW
        \raisebox{1.23\height}{\rotatebox[origin=c]{90}{\scriptsize \text{Ground truth}}} &
        \includegraphics[width=0.11\linewidth]{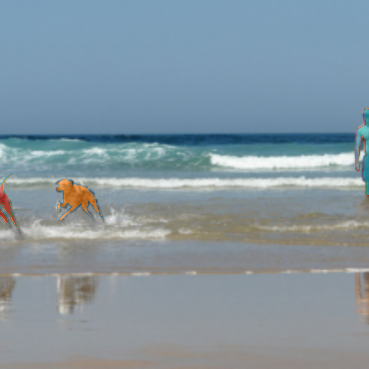} &
        \includegraphics[width=0.11\linewidth]{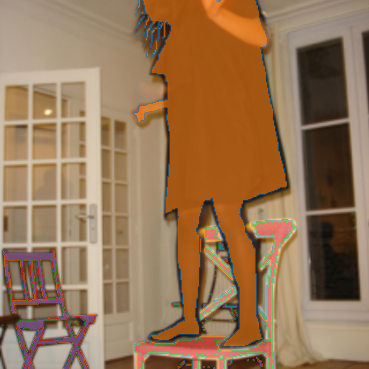} &
        \includegraphics[width=0.11\linewidth]{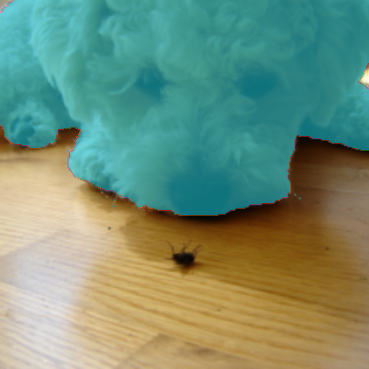} &
        \includegraphics[width=0.11\linewidth]{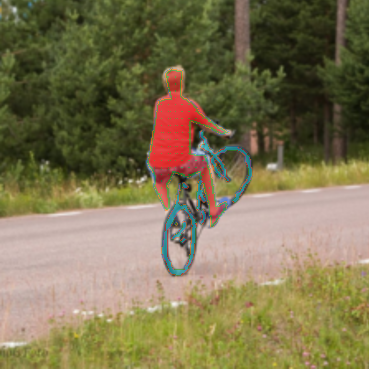} &
        \includegraphics[width=0.11\linewidth]{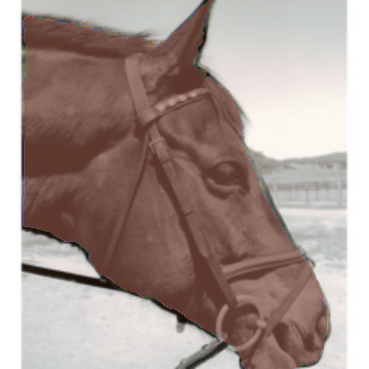} &
        \includegraphics[width=0.11\linewidth]{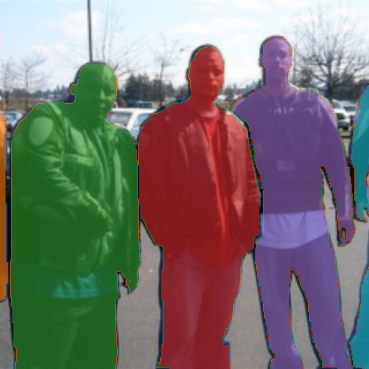} &
        \includegraphics[width=0.11\linewidth]{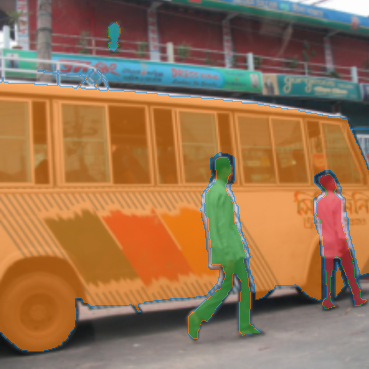} &
        \includegraphics[width=0.11\linewidth]{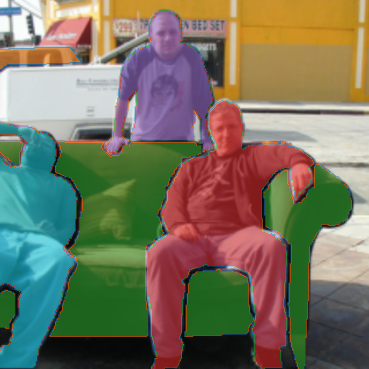} \\
    \end{tabular}
    \caption{\textbf{PASCAL VOC segmentation masks}. Images taken from PASCAL VOC, segmented by \ourspot \emph{(top row)}, SPOT \emph{(second row)}, \ourdino \emph{(third row)}, and DINOSAUR \emph{(fourth row)} compared against the ground truth \emph{(bottom row)}. For SPOT and DINOSAUR, segmentation masks derived from the decoder are shown, while for their respective \ourmethod variant, segmentation masks from the slot attention module are depicted.}
    \label{fig:appx_voc}
\end{figure*}

\begin{figure*}
    \centering
    \setlength{\tabcolsep}{2pt} % Adjust spacing between columns
    \begin{tabular}{ccccccccc} % First column for row labels, eight for images
        % MUFASA ROW
        \raisebox{1.7\height}{\rotatebox[origin=c]{90}{\scriptsize \text{\ourspot}}} &
        \includegraphics[width=0.11\linewidth]{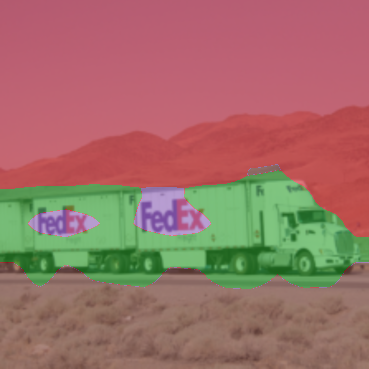} &
        \includegraphics[width=0.11\linewidth]{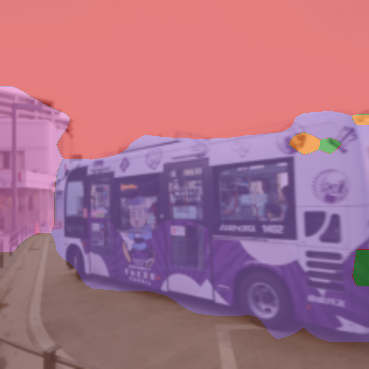} &
        \includegraphics[width=0.11\linewidth]{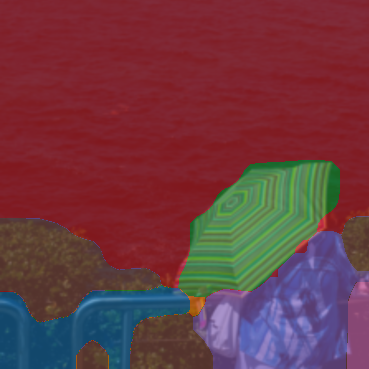} &
        \includegraphics[width=0.11\linewidth]{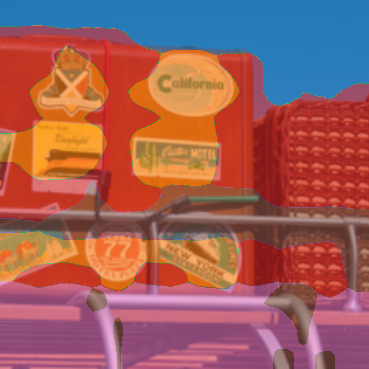} &
        \includegraphics[width=0.11\linewidth]{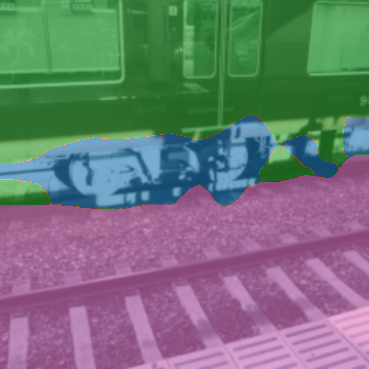} &
        \includegraphics[width=0.11\linewidth]{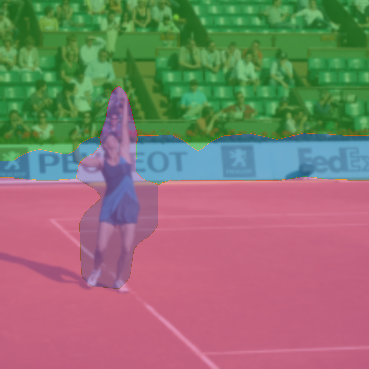} &
        \includegraphics[width=0.11\linewidth]{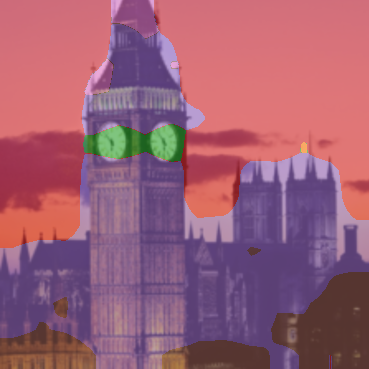} &
        \includegraphics[width=0.11\linewidth]{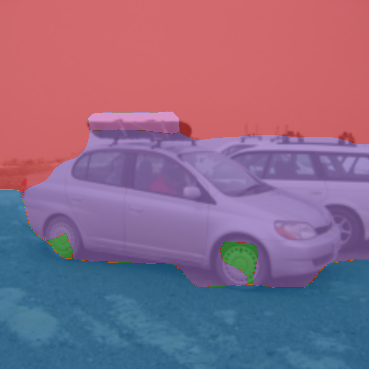} \\
        
        % SPOT ROW
        \raisebox{2.35\height}{\rotatebox[origin=c]{90}{\scriptsize \text{SPOT}}} &
        \includegraphics[width=0.11\linewidth]{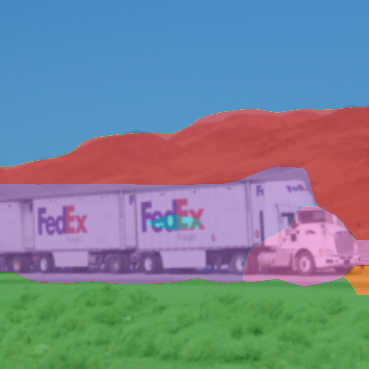} &
        \includegraphics[width=0.11\linewidth]{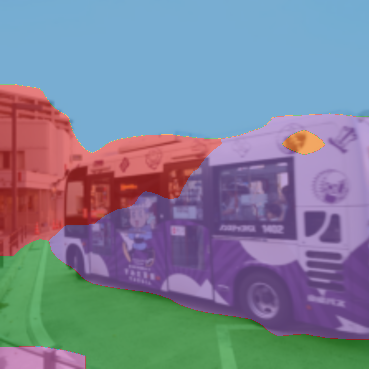} &
        \includegraphics[width=0.11\linewidth]{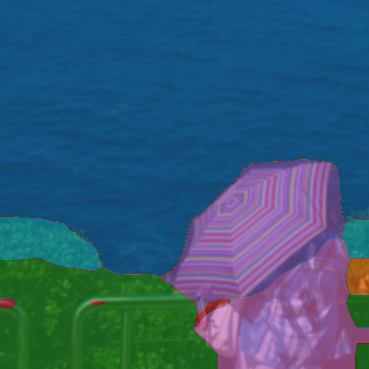} &
        \includegraphics[width=0.11\linewidth]{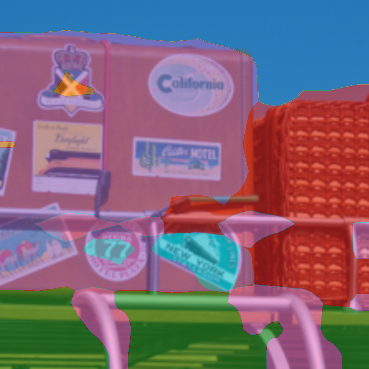} &
        \includegraphics[width=0.11\linewidth]{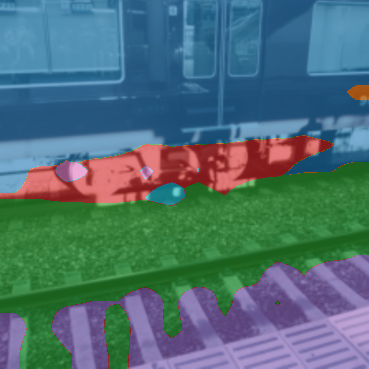} &
        \includegraphics[width=0.11\linewidth]{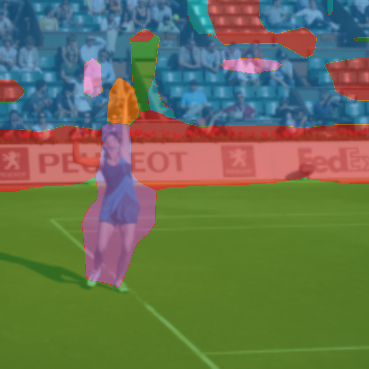} &
        \includegraphics[width=0.11\linewidth]{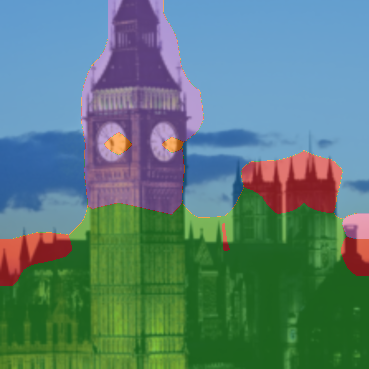} &
        \includegraphics[width=0.11\linewidth]{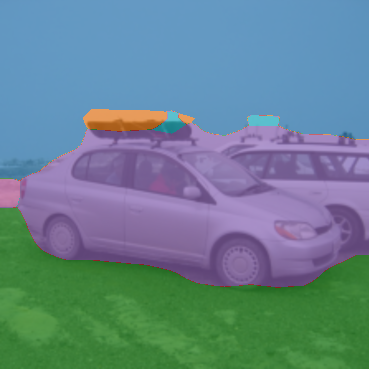} \\

        % MUFASA-D ROW
        \raisebox{1.05\height}{\rotatebox[origin=c]{90}{\scriptsize \text{\ourdino}}} &
        \includegraphics[width=0.11\linewidth]{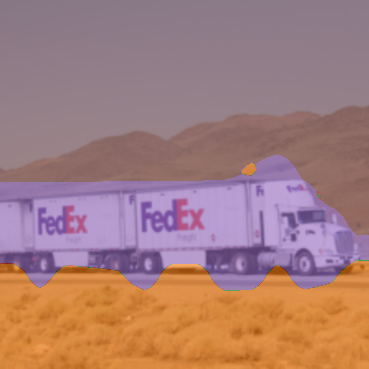} &
        \includegraphics[width=0.11\linewidth]{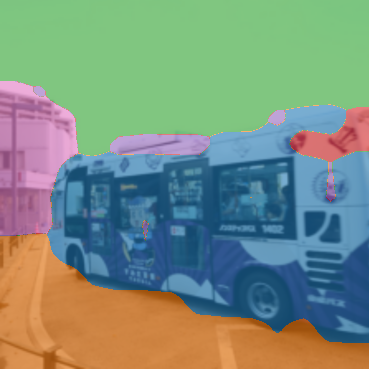} &
        \includegraphics[width=0.11\linewidth]{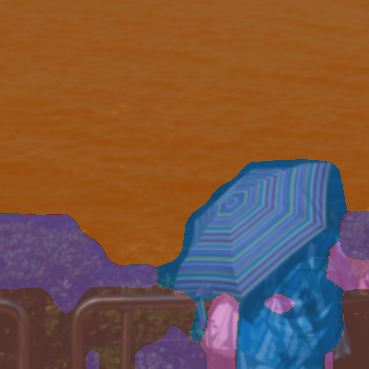} &
        \includegraphics[width=0.11\linewidth]{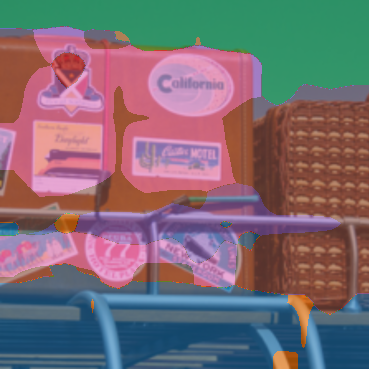} &
        \includegraphics[width=0.11\linewidth]{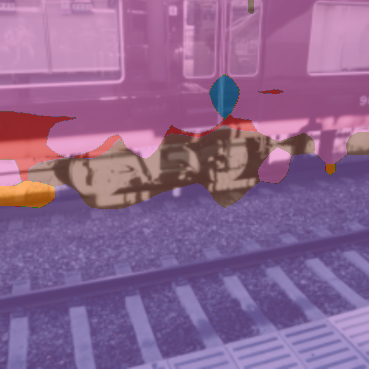} &
        \includegraphics[width=0.11\linewidth]{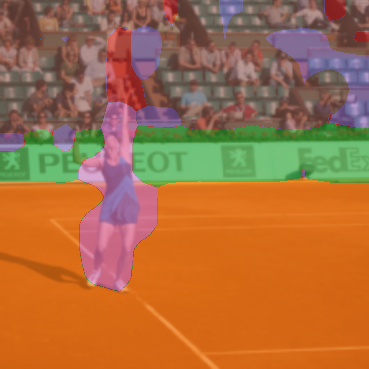} &
        \includegraphics[width=0.11\linewidth]{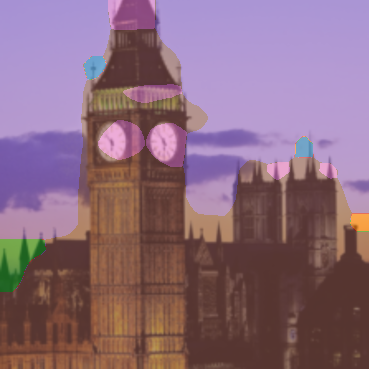} &
        \includegraphics[width=0.11\linewidth]{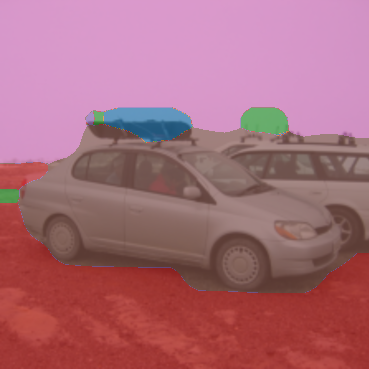} \\
        
        % DINOSUR ROW
        \raisebox{1.25\height}{\rotatebox[origin=c]{90}{\scriptsize \text{DINOSAUR}}} &
        \includegraphics[width=0.11\linewidth]{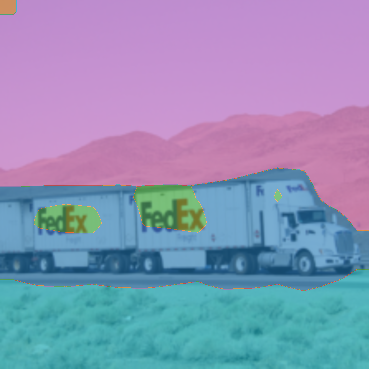} &
        \includegraphics[width=0.11\linewidth]{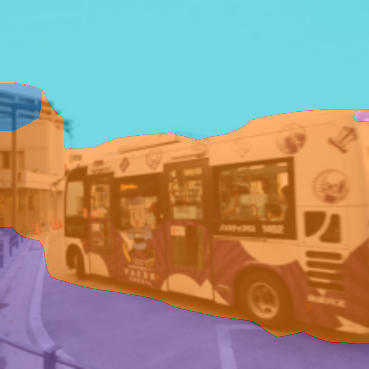} &
        \includegraphics[width=0.11\linewidth]{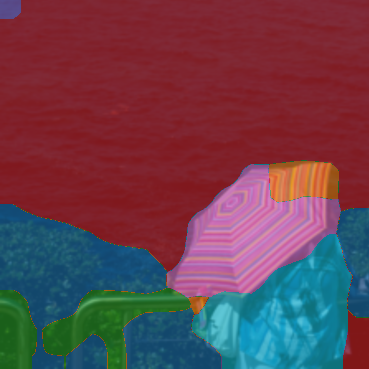} &
        \includegraphics[width=0.11\linewidth]{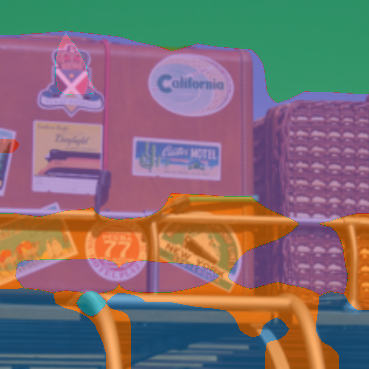} &
        \includegraphics[width=0.11\linewidth]{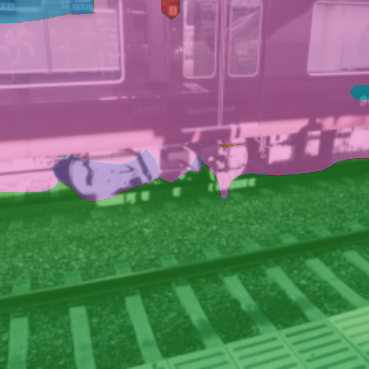} &
        \includegraphics[width=0.11\linewidth]{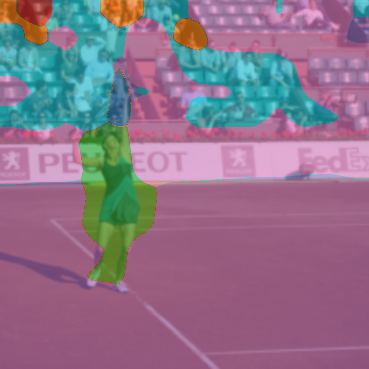} &
        \includegraphics[width=0.11\linewidth]{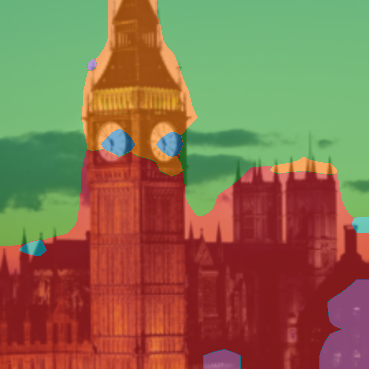} &
        \includegraphics[width=0.11\linewidth]{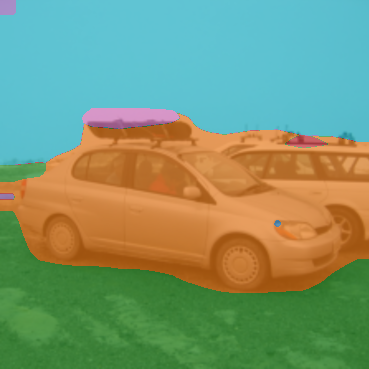} \\
        
        % GT ROW
        \raisebox{1.23\height}{\rotatebox[origin=c]{90}{\scriptsize \text{Ground truth}}} &
        \includegraphics[width=0.11\linewidth]{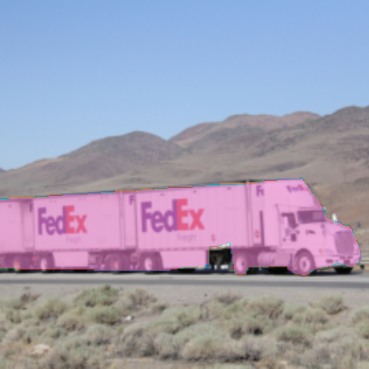} &
        \includegraphics[width=0.11\linewidth]{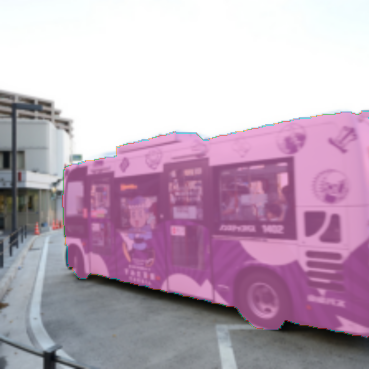 } &
        \includegraphics[width=0.11\linewidth]{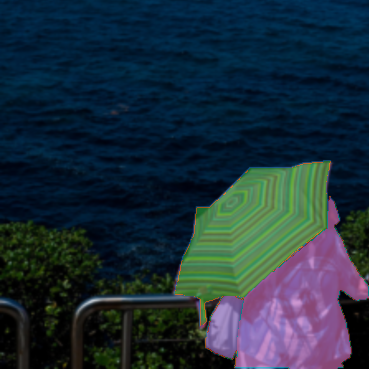 } &
        \includegraphics[width=0.11\linewidth]{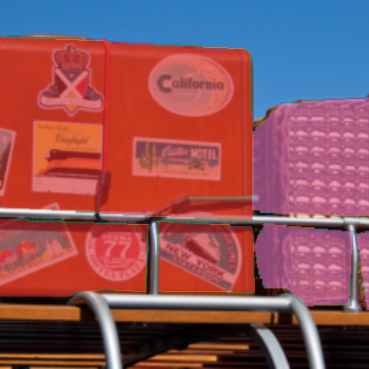 } &
        \includegraphics[width=0.11\linewidth]{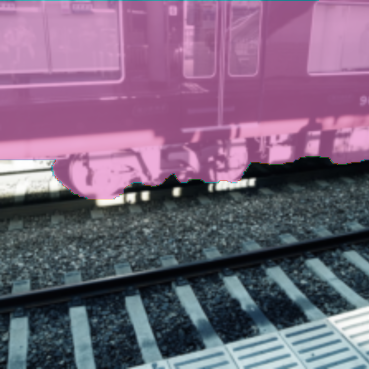} &
        \includegraphics[width=0.11\linewidth]{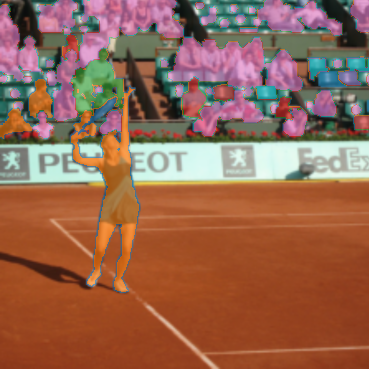} &
        \includegraphics[width=0.11\linewidth]{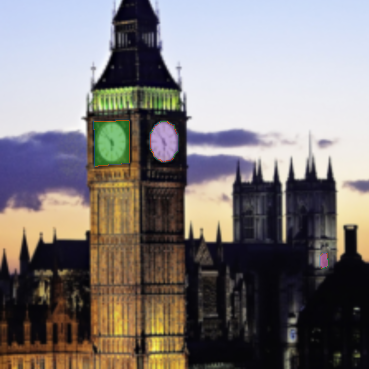} &
        \includegraphics[width=0.11\linewidth]{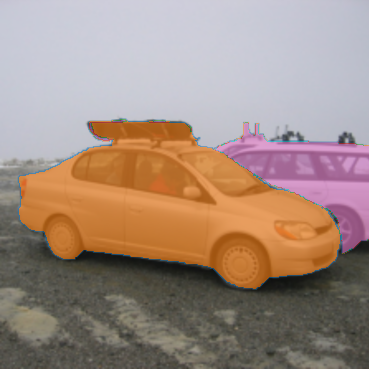} \\
    \end{tabular}
    \caption{\textbf{COCO segmentation masks.} Images taken from COCO, segmented by \ourspot \emph{(top row)}, SPOT \emph{(second row)}, \ourdino \emph{(third row)}, and DINOSAUR \emph{(fourth row)} compared against the ground truth \emph{(bottom row)}. For SPOT and DINOSAUR, segmentation masks derived from the decoder are shown, while for their respective \ourmethod variant, segmentation masks from the slot attention module are depicted.}
    \label{fig:appx_coco}
\end{figure*}

\begin{figure*}
    \centering
    \setlength{\tabcolsep}{2pt} % Adjust spacing between columns
    \begin{tabular}{ccccccccc} % First column for row labels, eight for images
        % MUFASA ROW
        \raisebox{1.7\height}{\rotatebox[origin=c]{90}{\scriptsize \text{\ourspot}}} &
        \includegraphics[width=0.11\linewidth]{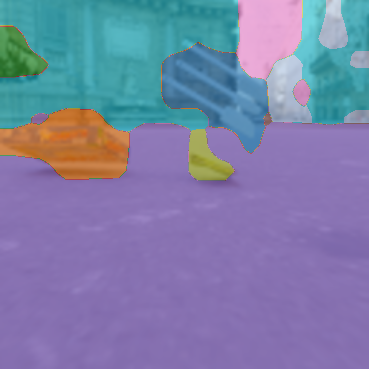} &
        \includegraphics[width=0.11\linewidth]{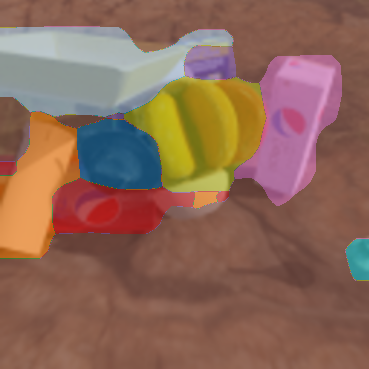} &
        \includegraphics[width=0.11\linewidth]{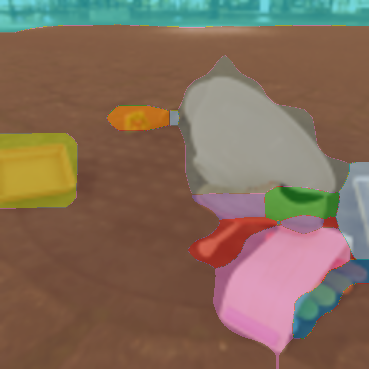} &
        \includegraphics[width=0.11\linewidth]{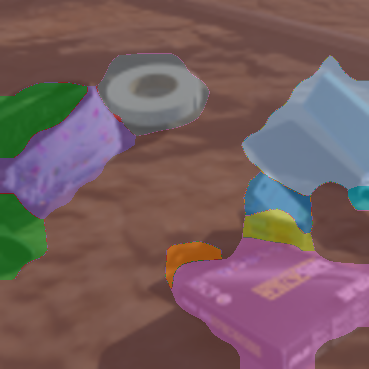} &
        \includegraphics[width=0.11\linewidth]{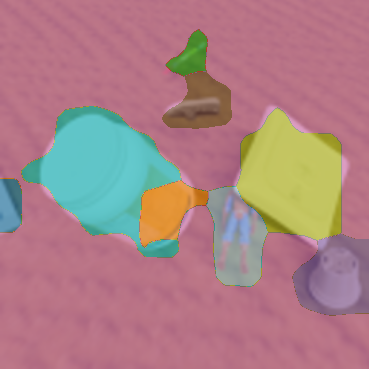} &
        \includegraphics[width=0.11\linewidth]{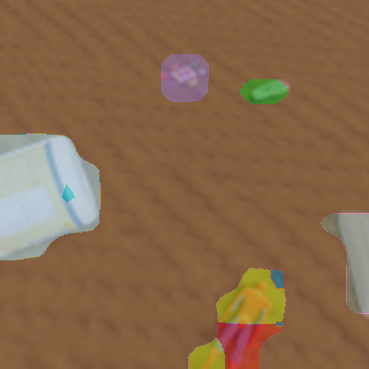} &
        \includegraphics[width=0.11\linewidth]{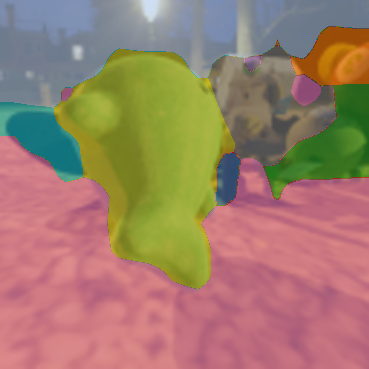} &
        \includegraphics[width=0.11\linewidth]{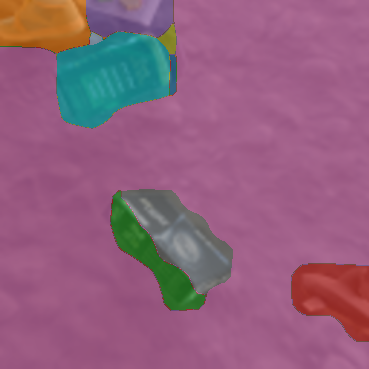} \\
        
        % SPOT ROW
       \raisebox{2.35\height}{\rotatebox[origin=c]{90}{\scriptsize \text{SPOT}}} &
        \includegraphics[width=0.11\linewidth]{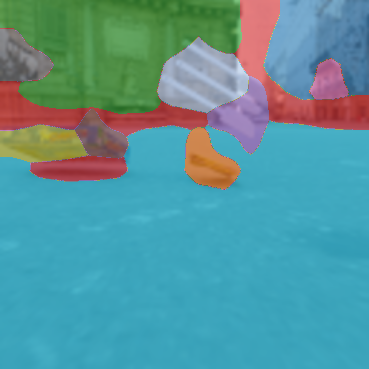} &
        \includegraphics[width=0.11\linewidth]{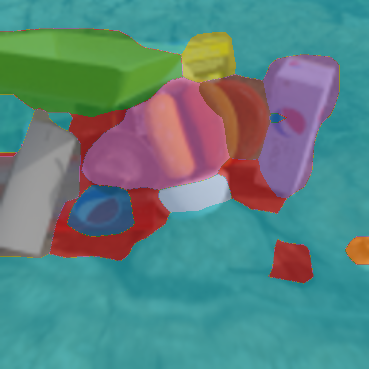} &
        \includegraphics[width=0.11\linewidth]{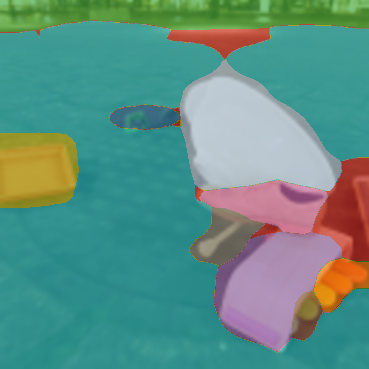} &
        \includegraphics[width=0.11\linewidth]{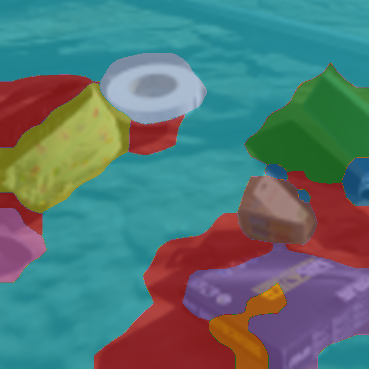} &
        \includegraphics[width=0.11\linewidth]{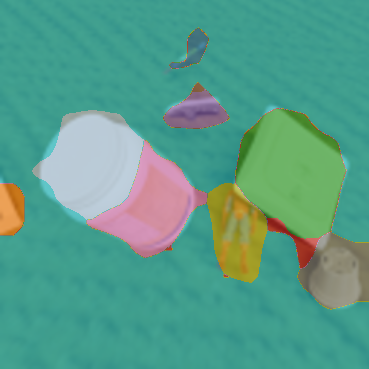} &
        \includegraphics[width=0.11\linewidth]{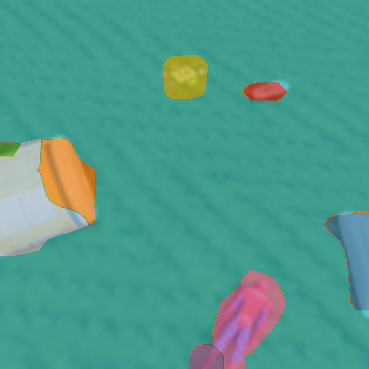} &
        \includegraphics[width=0.11\linewidth]{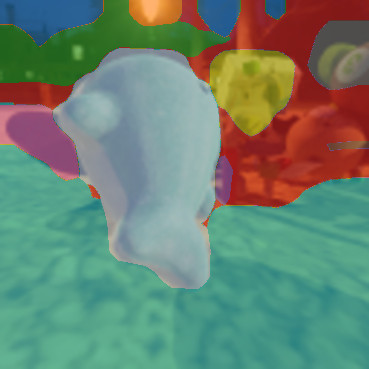} &
        \includegraphics[width=0.11\linewidth]{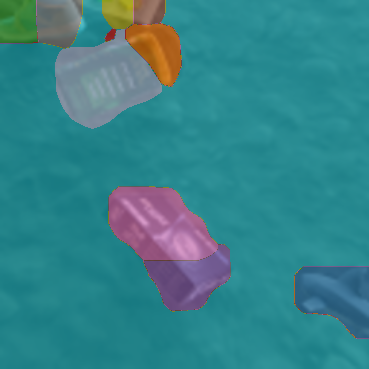} \\

        % MUFASA-D ROW
        \raisebox{1.05\height}{\rotatebox[origin=c]{90}{\scriptsize \text{\ourdino}}} &
        \includegraphics[width=0.11\linewidth]{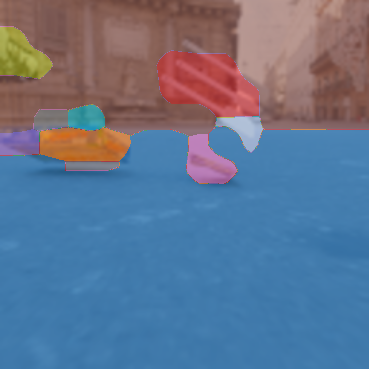} &
        \includegraphics[width=0.11\linewidth]{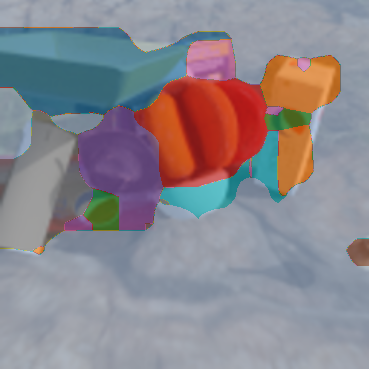} &
        \includegraphics[width=0.11\linewidth]{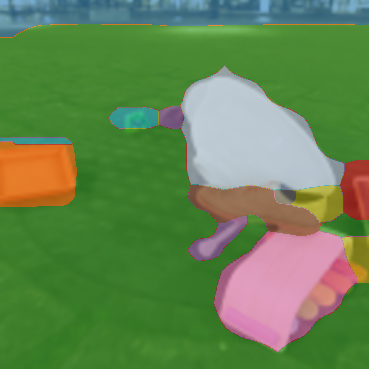} &
        \includegraphics[width=0.11\linewidth]{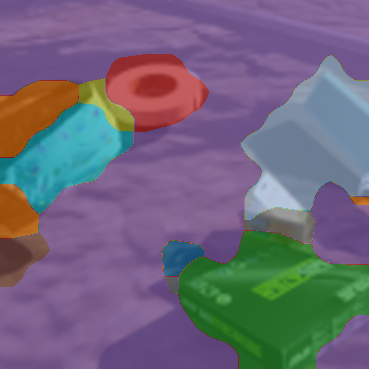} &
        \includegraphics[width=0.11\linewidth]{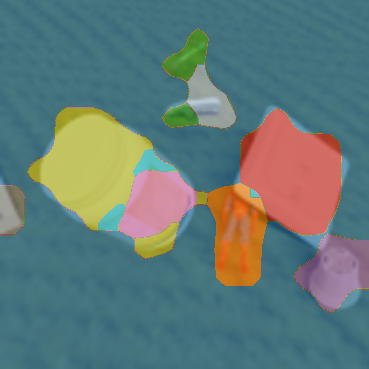} &
        \includegraphics[width=0.11\linewidth]{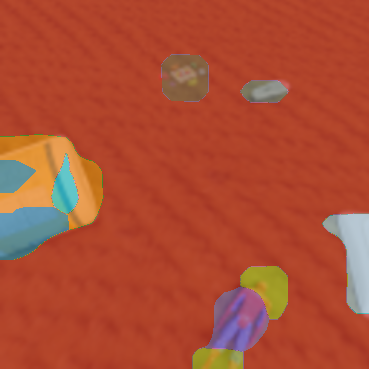} &
        \includegraphics[width=0.11\linewidth]{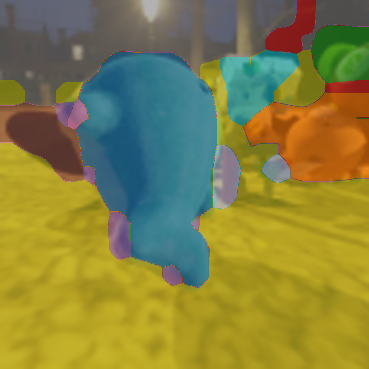} &
        \includegraphics[width=0.11\linewidth]{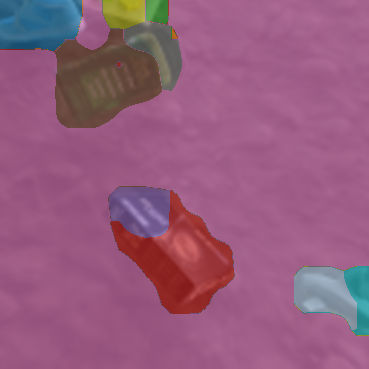} \\
        
        % DINOSUR ROW
        \raisebox{1.25\height}{\rotatebox[origin=c]{90}{\scriptsize \text{DINOSAUR}}} &
        \includegraphics[width=0.11\linewidth]{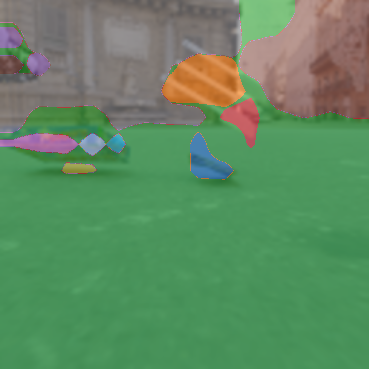} &
        \includegraphics[width=0.11\linewidth]{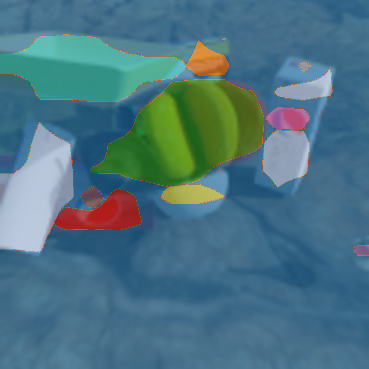} &
        \includegraphics[width=0.11\linewidth]{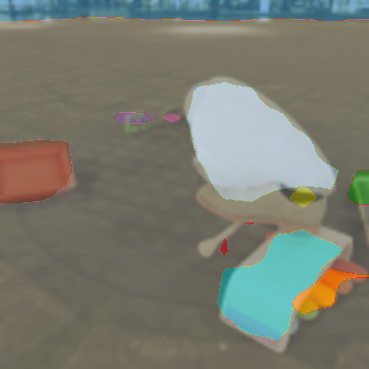} &
        \includegraphics[width=0.11\linewidth]{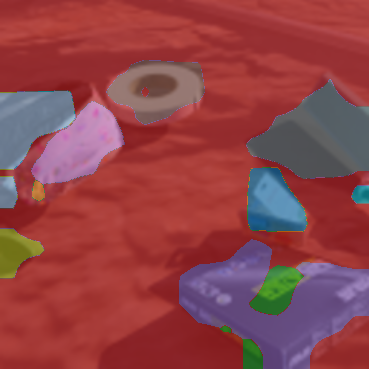} &
        \includegraphics[width=0.11\linewidth]{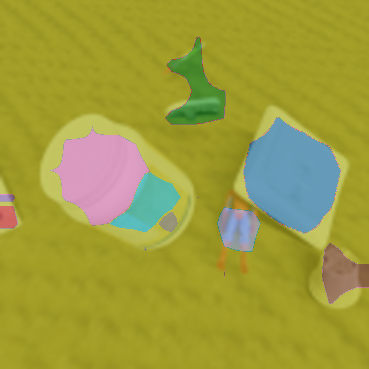} &
        \includegraphics[width=0.11\linewidth]{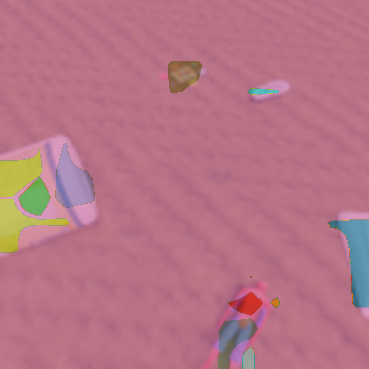} &
        \includegraphics[width=0.11\linewidth]{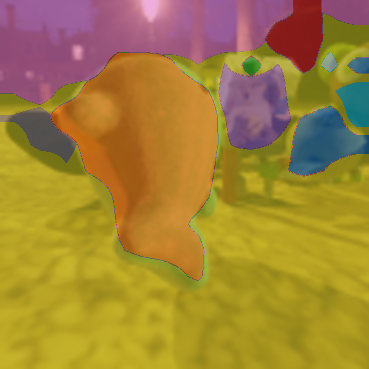} &
        \includegraphics[width=0.11\linewidth]{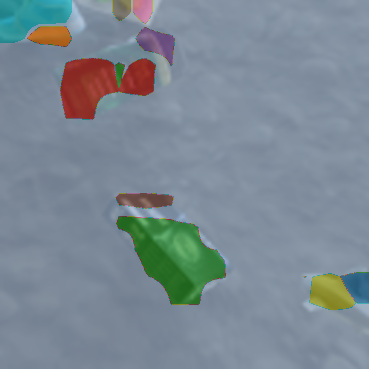} \\
        
        % GT ROW
        \raisebox{1.23\height}{\rotatebox[origin=c]{90}{\scriptsize \text{Ground truth}}} &
        \includegraphics[width=0.11\linewidth]{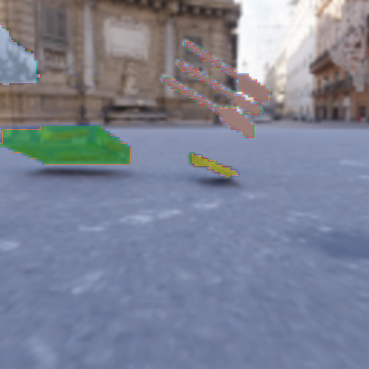} &
        \includegraphics[width=0.11\linewidth]{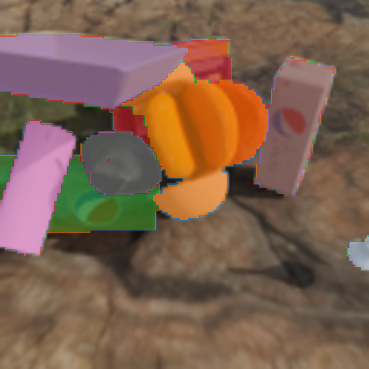} &
        \includegraphics[width=0.11\linewidth]{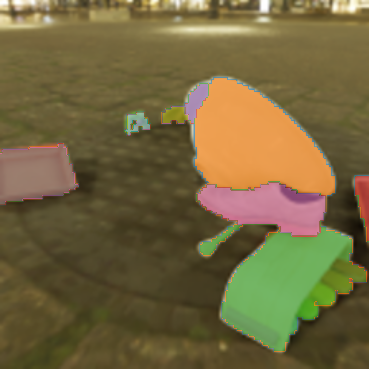} &
        \includegraphics[width=0.11\linewidth]{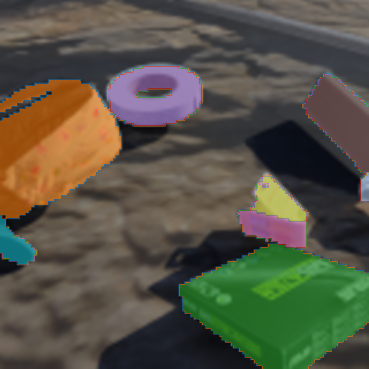} &
        \includegraphics[width=0.11\linewidth]{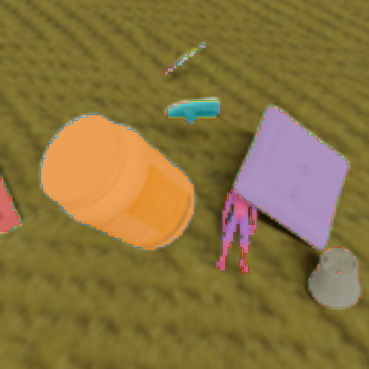} &
        \includegraphics[width=0.11\linewidth]{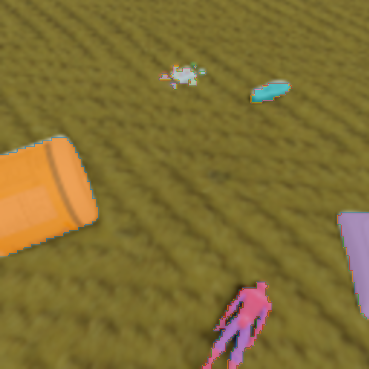} &
        \includegraphics[width=0.11\linewidth]{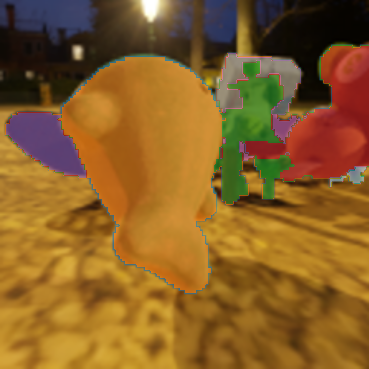} &
        \includegraphics[width=0.11\linewidth]{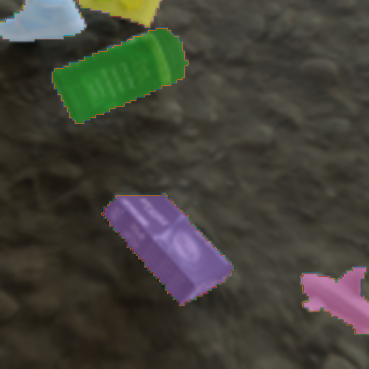} \\
    \end{tabular}
    \caption{\textbf{MOVi-C segmentation masks}. Images taken from MOVi-C, segmented by \ourspot \emph{(top row)}, SPOT \emph{(second row)}, \ourdino \emph{(third row)}, and DINOSAUR \emph{(fourth row)} compared against the ground truth \emph{(bottom row)}. For SPOT and DINOSAUR, segmentation masks derived from the decoder are shown, while for their respective \ourmethod variant, segmentation masks from the slot attention module are depicted.}
    \label{fig:appx_movic}
\end{figure*}

We provide further segmentation masks for \ourdino and \ourspot compared against their respective baselines, as well as the ground truths for PASCAL VOC in \cref{fig:appx_voc}, COCO in \cref{fig:appx_coco}, and MOVi-C in \cref{fig:appx_movic}. They provide an extended overview over different settings and motives, such as close-up objects, landscapes, or a composition of multiple small objects to emphasize \ourmethod's abilities to decompose various kinds of scenes into meaningful entities.


\end{document}